\documentclass[journal]{IEEEtran}
\ifCLASSINFOpdf
  \usepackage[pdftex]{graphicx}
\else
\fi
\usepackage{amsmath}

\DeclareMathOperator{\sign}{sign}
\DeclareMathOperator{\KL}{KL}
\usepackage{dblfloatfix}
\usepackage{url}

\usepackage{xcolor}

\usepackage{enumerate}
\usepackage{xspace}

\newcommand{\BASELINE}{{\footnotesize \textsf{BASELINE}}\xspace}
\newcommand{\VOTINGBASELINE}{{\footnotesize \textsf{VOTING\_BASELINE}}\xspace}
\newcommand{\ROBADVGEN}{{\footnotesize \textsf{ROBUST\_ADV\_GEN}}\xspace}
\newcommand{\DEFENSEGEN}{{\footnotesize \textsf{DEFENSE\_GEN}}\xspace}
\newcommand{\ENHANCED}{{\footnotesize \textsf{ENHANCED}}\xspace}
\newcommand{\VOTINGENHANCED}{{\footnotesize \textsf{VOTING\_ENHANCED}}\xspace}

\newcommand{\FIGSIZE}{3.0in}

\newcommand{\FIGSIZEXL}{3.5in}

\begin{document}
%
\title{Detecting Adversarial Examples by Input Transformations, Defense Perturbations, and Voting\thanks{\textcopyright 2021 IEEE.  Personal use of this material is permitted.  Permission from IEEE must be obtained for all other uses, in any current or future media, including reprinting/republishing this material for advertising or promotional purposes, creating new collective works, for resale or redistribution to servers or lists, or reuse of any copyrighted component of this work in other works.}}
%
%
%
\author{Federico Nesti,
		Alessandro Biondi,~\IEEEmembership{Member,~IEEE,} and
        Giorgio Buttazzo,~\IEEEmembership{Fellow,~IEEE}
\\
\IEEEauthorblockA{Department of Excellence in Robotics \& AI, Scuola Superiore Sant'Anna, Pisa, Italy}
\\
\IEEEauthorblockA{Tecip Institute, Scuola Superiore Sant'Anna, Pisa, Italy}
}

\maketitle

\begin{abstract}
Over the last few years, convolutional neural networks (CNNs) have proved to reach super-human performance in visual recognition tasks. However, CNNs can easily be fooled by adversarial examples, i.e., maliciously-crafted images that force the networks to predict an incorrect output while being extremely similar to those for which a correct output is predicted. Regular adversarial examples are not robust to input image transformations, which can then be used to detect whether an adversarial example is presented to the network. Nevertheless, it is still possible to generate adversarial examples that are robust to such transformations.

This paper extensively explores the detection of adversarial examples via image transformations and proposes a novel methodology, called \textit{defense perturbation}, to detect robust adversarial examples with the same input transformations the adversarial examples are robust to. Such a \textit{defense perturbation} is shown to be an effective counter-measure to robust adversarial examples.

Furthermore, multi-network adversarial examples are introduced. This kind of adversarial examples can be used to simultaneously fool multiple networks, which is critical in systems that use network redundancy, such as those based on architectures with majority voting over multiple CNNs. An extensive set of experiments based on state-of-the-art CNNs trained on the Imagenet dataset is finally reported.
\end{abstract}


%
\IEEEpeerreviewmaketitle



\section{Introduction} \label{s:intro}

During the last few years, convolutional neural networks (CNNs) have been used in many fields with outstanding, and sometimes super-human, performance \cite{Khan_2020}, \cite{mishkin}, \cite{mnih}. At the same time, a lot of research has been devoted to the robustness of such models, often focusing on adversarial examples \cite{szegedy2014}, \cite{ZhangTransactions}.

Adversarial examples (AEs) are maliciously-crafted inputs (in this case, images) that have the power to fool a neural network by forcing its prediction towards an erroneous class, by slightly changing the intensity values of the pixels, keeping almost the same digital representation. From the perspective of a human observer, a typical AE has the same visual appearance as the original image.

AEs are a serious concern for the safety and security of systems based on artificial intelligence (AI).
For instance, they could be used to attack a neural network for object recognition in an autonomous (or even semiautonomous) vehicle, possibly causing a catastrophic consequence \cite{bojarski2016end}, \cite{kurakin2016adversarial}, \cite{nassi}, \cite{survey}.

\begin{figure}[!t]
\centering
\includegraphics[width=\FIGSIZEXL]{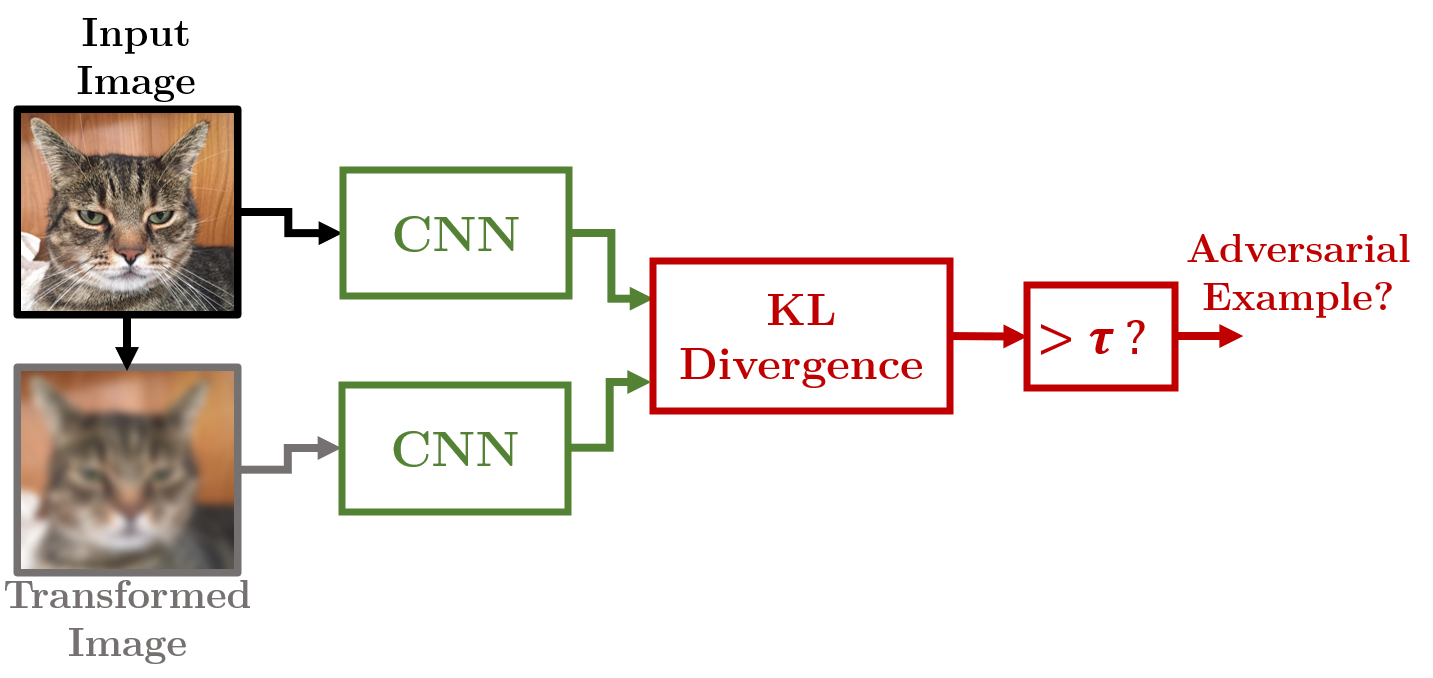}
\caption{\BASELINE architecture used to detect standard adversarial examples.}
\label{f:single_net_detection}
\end{figure}

These facts motivate the search for an effective defense against AEs.
For instance, Guo et al. \cite{guo2017countering} showed that standard\footnote{In this paper, standard AEs refer to those AEs crafted without considering image transformations, using the classical formulation presented in Section \ref{s:model} by Equation \eqref{e:min_adv}.} AEs
are not robust to input transformations, such as translation, rotation, and other input changes. These findings suggest that standard AEs can be detected by measuring how the network prediction changes when an input image is replaced with the same one processed with a given transformation.
The two network predictions can then be compared by the Kullback-Leibler (KL) divergence \cite{visionguard}, so that an input image is considered to be adversarial if the two predictions are ``distant" from each other, and non-adversarial if the two predictions are ``close" to each other.
A binary classification can then be obtained by applying a threshold to the output of the KL module. This approach can easily be implemented using the architecture illustrated in Figure~\ref{f:single_net_detection}, referred to as BASELINE detection architecture.

Input transformations are attractive because they are simple, require a limited computational cost (thus can be performed at run-time), and do not need any training procedure. Nevertheless, they suffer from two major problems: (i) they might not have a good detection performance (also due to the accuracy degradation they may cause), and (ii) it has been shown~\cite{robust} that it is still possible to construct AEs that are \emph{robust} to input transformations.

\subsection{This paper}

This work addresses the two problems mentioned above. To evaluate how different input transformations affect the performance of the detection system and the accuracy of the network, an extensive experimental campaign is reported both for adversarial and non-adversarial images. To the best of our knowledge, a similar experimental study has never been presented in the literature.

To cope with the second problem, different new methods and architectures are proposed to include a novel counter-measure for detecting robust AEs. The counter-measure is based on the assumption that the defender is aware that the attacker knows how to craft robust AEs and has also knowledge about the transformations that are used. To counteract this kind of attacks, a new defense method, called \emph{defense perturbation}, is introduced to ``convert" robust AEs into non-robust ones. Such a defense perturbation is generated from robust AEs, similarly to how universal adversarial perturbations \cite{universal} are generated.

As a further evolution, this work considers a multi-network architecture composed of three different state-of-the-art CNNs for image recognition (trained on ImageNet \cite{ILSVRC15}), which are combined by means of a majority voting algorithm (2 out of 3). For instance, this approach was proposed by Biondi et al.~\cite{biondi2019safe} as a solution for adopting neural networks in safety-critical control systems. 
Although there exist methods to transfer AEs between different models and architectures \cite{xie2018improving}, \emph{multi-network adversarial examples} are introduced as a new kind of AEs that are capable of fooling multiple networks simultaneously without applying any network-specific perturbation.

\smallskip
\noindent
\textbf{Contribution and paper structure.}
In summary, this paper makes the following novel contributions:
\begin{itemize}
  \item It reports an extensive experimental evaluation on the detection capabilities of input transformations. 
  \item It presents a methodology for setting up a counter-measure against robust adversarial examples.
  \item It introduces multi-network adversarial examples to systematically fool multiple networks at once. 
  \item The proposed methods are finally combined to design an effective architecture for detecting robust adversarial examples.
\end{itemize}

The rest of the paper is organized as follows: Section \ref{s:model} introduces the problem and the notation; Section \ref{s:bkg} provides a brief overview of background and related work; Section \ref{s:proposed} describes the proposed approaches; Section \ref{s:exp} presents the experimental results; and Section \ref{s:concl} states the conclusions.

\section{System and Threat model} \label{s:model}

This paper considers CNNs for image recognition. Let $\mathcal{X} = [0, 1]^{h\times w \times c}$ be the image space (of dimensions $w$, $h$, and $c$, namely the width, height, and the number of channels of the image, respectively).
A CNN behaves as a function $f(\cdot): \mathcal{X} \rightarrow [0, 1]^n$ that takes as input an image of fixed dimensions and outputs a discrete probability distribution vector with dimension $n$ equal to the number of classes considered for the classification problem. The class predicted by a neural network classifier is represented by the function $\hat{f}(\cdot) = \arg\max f(\cdot)$.

An adversarial perturbation can be modeled as a tensor $r\in [-\epsilon, \epsilon]^{h\times w \times c}$, where $\epsilon \in (0, 1)$ (typically small) is a parameter typically named \emph{adversarial strength}. Given a source image $x\in \mathcal{X}$, an AE is then an image $x+r \in \mathcal{X}$ such that, if $\hat{f}(x)=t$, where $t$ is the correct target class of the image $x$, then $\hat{f}(x+r)=t_{adv}$, being $t_{adv} \neq t$ the adversarial target class.

AEs can be characterized by (i) the \emph{knowledge level} of the attacker, (ii) the \emph{target specificity}, and (iii) the \emph{similarity metric} used to minimize the distance of an AE from the source image.
The AEs considered in this paper are:
\begin{enumerate}[(i)]
  \item \textbf{White-box}, i.e., the attacker has perfect knowledge of the structure of the network and its parameters (this is the strongest type of attack);
  \item \textbf{Targeted}, i.e., they are crafted to force the prediction of the network to a specific class;
  \item Generated using the \textbf{$L_2$ norm}, i.e., the Euclidean norm, to express the distance of an AE from the source image.
\end{enumerate}

An AE of this type can be crafted by optimizing
\begin{equation} \label{e:min_adv}
\min_{r} [\mathcal{L}(f(x+r), t_{adv}) + k \|r\|_2^2],
\end{equation}

\noindent where $\mathcal{L}$ in the above equation is a loss function expressing the distance between the target $t_{adv}$ and the output distribution of the network, and $k$ is a constant
that reflects how much the magnitude of the perturbation is weighted in the minimization. Typically, the loss function $\mathcal{L}$ is a cross-entropy, but it can also assume other more complex forms (e.g., as in the Carlini-Wagner attack~\cite{carlini}, described in Section \ref{s:bkg}). The optimization is iteratively performed with a stochastic gradient descent approach for a certain number of epochs: this optimization strategy is used throughout the whole paper for the generation of AEs.
When searching for the minimum-perturbation AE, a further optimization is usually performed to find the optimal $k$. This fine-grained optimization is out of the scope of this work, and hence $k$ is fixed to 0.01 for all the AEs.

\begin{figure}[!t]
\centering
\includegraphics[width=\FIGSIZEXL]{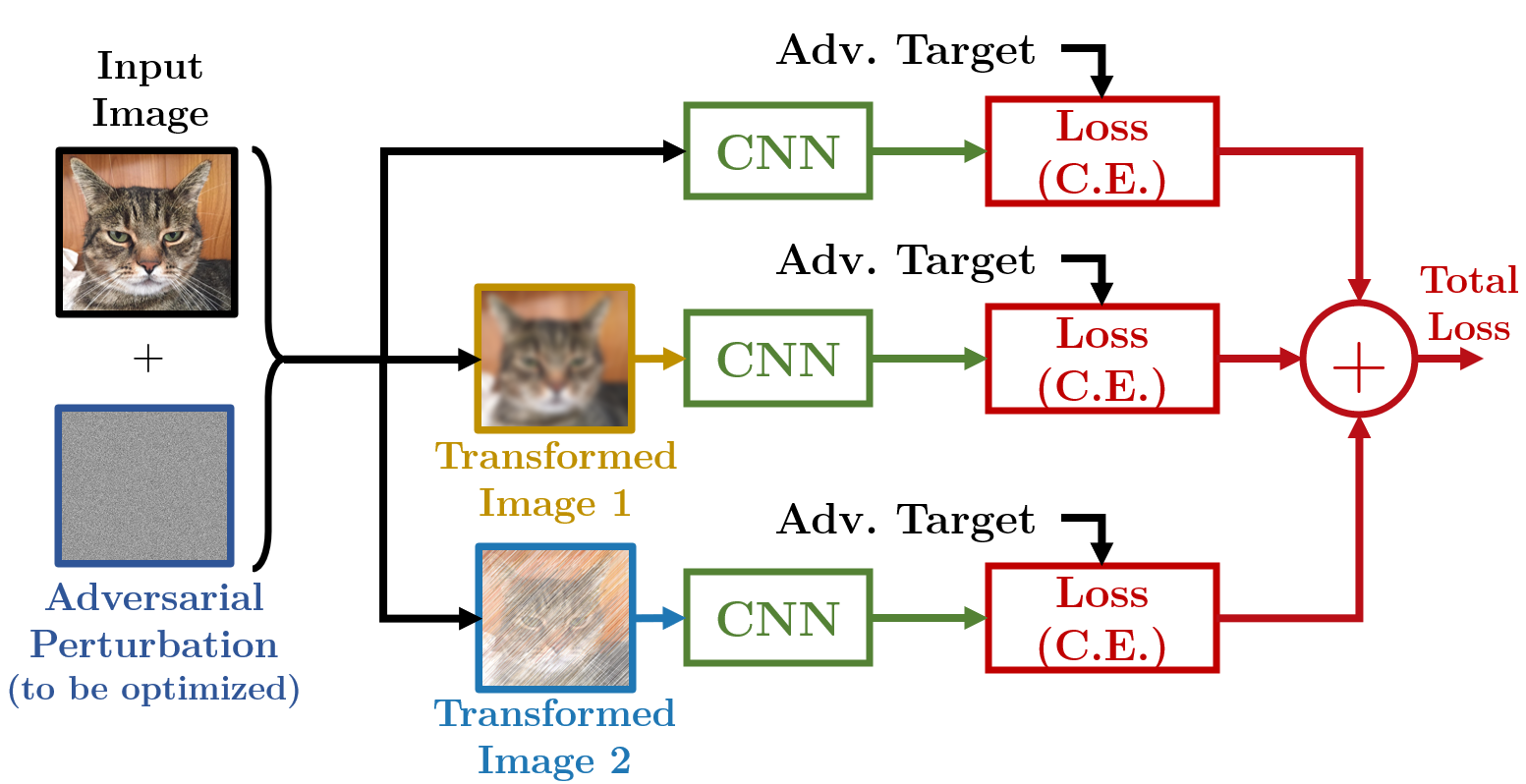}
\caption{\ROBADVGEN architecture used to craft robust adversarial examples.}
\label{f:robust_adv}
\end{figure}

The AEs crafted with the procedure described above are sensitive to input transformations and are referred to as \textbf{standard} AEs.
Conversely, \textbf{robust} AEs are generated with a slightly modified optimization process based on the architecture illustrated in Figure \ref{f:robust_adv}, which is named \ROBADVGEN. Let $\{ g_j(x; \theta_j), j=1, \dots N\}$ be a set of $N$ transformations of the input image (e.g., translation, rotation, etc.), each of which depends on a parameter $\theta_j$. AEs that are robust to each of these transformations can then be generated by minimizing

\begin{equation} \label{e:min_robust}
\min_{r} [\mathcal{L}(f(x+r), t_{adv}) + \sum\limits_{j=1}^N \mathcal{L}(f(g_j(x+r; \theta_j)), t_{adv}) + k \|r\|_2^2].
\end{equation}

Most image transformations depend on a parameter $\theta_j$ that varies in a known range. At each iteration step of the optimization procedure, the parameters of the transformations are uniformly sampled from their corresponding ranges,
hence generating AEs that are robust to a wide spectrum of configurations of the image transformations.
Details on the generation of robust AEs used in this work are reported in Section \ref{s:exp}.

\begin{figure}[!t]
\centering
\includegraphics[width=\FIGSIZEXL]{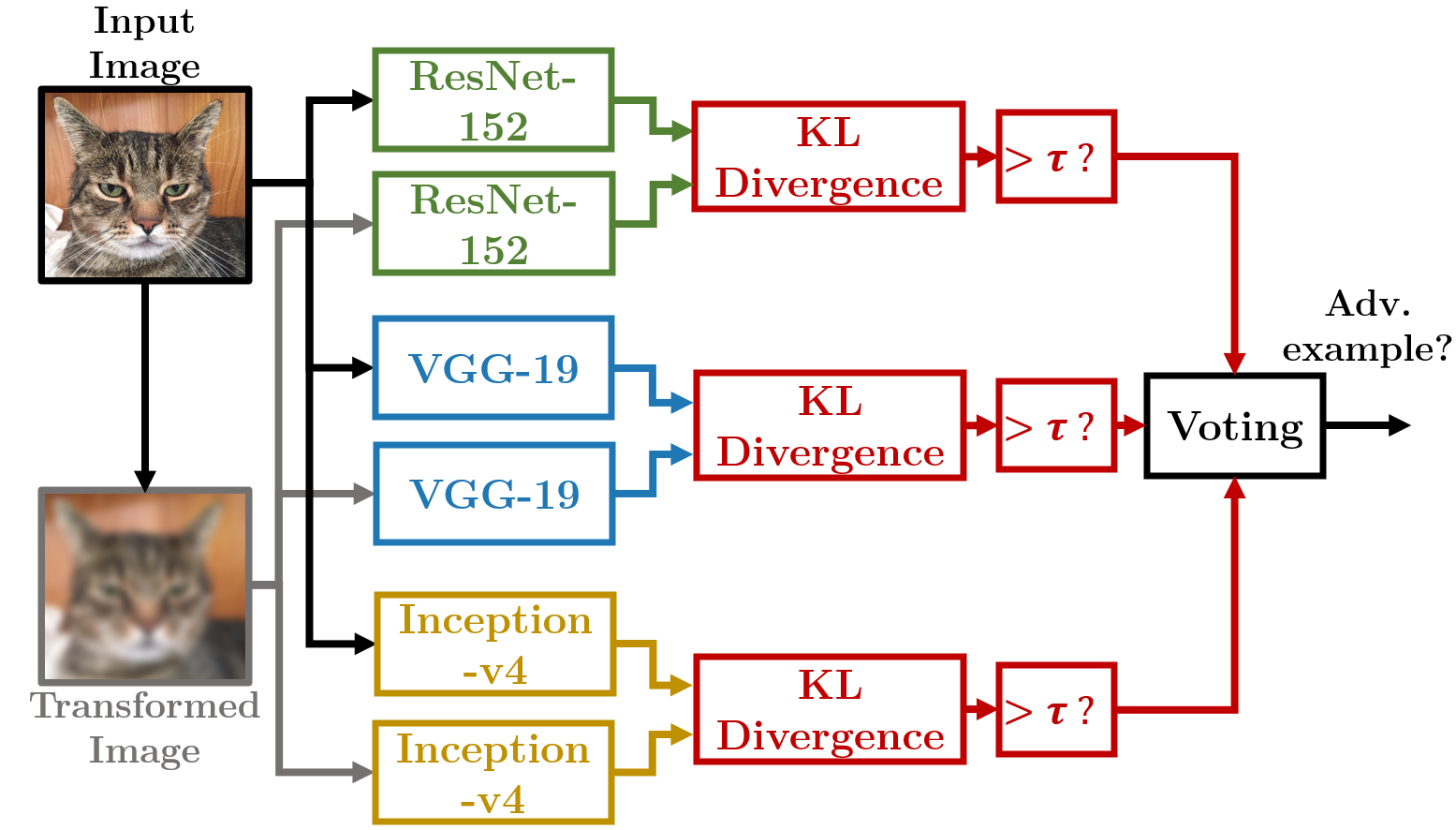}
\caption{The \VOTINGBASELINE architecture used in this work.}
\label{f:arch1}
\end{figure}

Since a deep neural network is naturally prone to classification errors, especially when considering grey cases (i.e., previously unseen, possibly harmful inputs), an ensemble of networks can be used~\cite{biondi2019safe} to mitigate the errors of a single network, by combining the outputs with a voting algorithm.
The resulting architecture is illustrated in Figure \ref{f:arch1} and consists in applying a voting algorithm (such as majority voting) over $M$ different CNNs. This architecture is referred to as \VOTINGBASELINE. This paper only considers $M=3$ networks, which are described in details in Section~\ref{s:exp}.
Standard AEs, as those described above, result adversarial for a single network. With this kind of voting architecture, an AE that fools a single network is a less dangerous threat, since the voting algorithm will cover for that mistake with the predictions from the other two networks.

Note that single-network AEs may also fool other networks. However, in order to assess the detection capabilities of such system, it is more convenient to craft images that are adversarial for the three networks simultaneously. Such AEs are also adversarial for each single network and can be used to evaluate the detection performance of input transformations for each single network.
For this reason, multi-network AEs are introduced. Given a set of $M$ classifiers, each denoted by $f_i(\cdot), \; i=1, \ldots, M$,
and a set of $N$ transformations of the input image, denoted as $\{ g_j(x; \theta_j), j=1, \dots N\}$,
%
%
\emph{multi-network robust adversarial examples} can be crafted by minimizing the following extension of Equation \eqref{e:min_robust}:
\begin{multline}
\min_{r} [\sum\limits_{i=1}^M \mathcal{L}(f_i(x+r), t_{adv})+ \\+ \sum\limits_{j=1}^N\sum\limits_{i=1}^M \mathcal{L}(f_i(g_j(x+r; \theta)), t_{adv}) + k \|r\|_2^2].
\end{multline}
If generated in this way, an image results to be adversarial not only for all the $M$ networks, but also for all the $N$ transformations.

These AEs have been used for the experimental evaluation and proved to be more difficult to detect than single-networks AEs, as they are adversarial for each of the selected networks. The methodology presented above extends a series of state-of-the-art attack methods, which are reviewed next.

\section{Background and Related work} \label{s:bkg}
The literature concerning AEs has been growing exponentially over the last years, and several different attack and defense methods have been proposed. This section reviews the most common attacks and defenses, highlighting how this work is positioned within the published literature.

\subsection{Attacks}
\paragraph{L-BFGS attack} Szegedy et al. \cite{szegedy2014} first introduced AEs against deep neural networks. The approach is the one described in Equation~\eqref{e:min_adv}. The loss function they used is a cross-entropy.
\paragraph{Carlini-Wagner (CW) L2 attack} Carlini and Wagner \cite{carlini} proposed a set of more complex loss functions (and different threat models),
among which the most relevant is $\mathcal{L} = \max(0, \max\{Z(x+r)_k: k\neq t_{adv}\}-Z(x+r)_{t_{adv}})$, where $Z(x)$ is the output of the logits layer (i.e., before the softmax layer), and $Z(x)_k$ is the logit value corresponding to class $k$.
The effect of using this loss is that, during the optimization process, this term falls to zero as soon as the predicted class matches the adversarial target, without considering the confidence associated with that prediction. This allows minimizing the adversarial perturbation only 
as soon as the first term drops to zero.

\paragraph{FGSM attack} Fast Gradient Sign Method was introduced by Goodfellow et al. \cite{goodfellow2014explaining}, and it is a non-iterative, untargeted method for adverarsarial examples generation. It is sufficient to compute the gradient with respect to the image of a certain loss function $\nabla_x \mathcal{L}(f(x), t)$ (usually cross-entropy).
The adversarial perturbation is then generated as $\epsilon \sign (\nabla_x \mathcal{L}(f(x), t))$.
Although this type of AE is one of the most popular for its simplicity and fast generation, the focus of this paper is on AEs that are able to simultaneously fool multiple networks, and targeted AEs are more suited for this purpose. Hence, this kind of attack is not considered in this work since, for an accurate assessment of the detection capabilities of this multi-network system, it would require an accurate filtering of the perturbations that resulted adversarial for the ensemble of the three networks.

\paragraph{Robust adversarial examples} AEs can be made robust (in expectation) to a certain transformation distribution $T$ \cite{robust} by randomizing the parameters of the transformation during the optimization of the adversarial perturbation. This kind of AEs are, by construction, not easily detectable with the
\BASELINE architecture (Figure \ref{f:single_net_detection}).

\paragraph{Universal adversarial perturbation} this kind of attack crafts an image-agnostic adversarial perturbation \cite{universal}, which is generated to result adversarial for a set of different images belonging to different classes. A universal perturbation has the property of making regular images become AEs, even when considering images that were not used for the optimization of the perturbation. Although this kind of attack is not considered in this work because weaker than the ones presented above, it is worth citing it since it inspired the defense perturbation presented in this paper, as described in Section \ref{s:proposed}.

\paragraph{Other attacks} The literature presents many other different attacks that are not considered in this paper for space limitations. Among these, the most relevant to us are DeepFool~\cite{deepfool} and JMSA~\cite{JMSA}.

\subsection{Defenses}
The defenses proposed in the literature can be divided into three categories, briefly described below:
\paragraph{Modifying Data} the defenses that fall in this category are the ones that modify the input data at run-time to defend from AEs. Possible approaches are based on \emph{data compression} and filtering \cite{guo2017countering}, \cite{prakash}, \cite{visionguard} or \emph{data randomization} \cite{xie2017mitigating}.
The underlying idea is similar to the one presented in this work. However, the objective of this work is to detect AEs rather than predicting the correct class; moreover, it presents an extensive experimental evaluation of the most common image transformations.

\paragraph{Modifying Model} this type of defenses includes the ones that act on the classifier in order to prevent attacks. For instance, \emph{regularization} (i.e., adding a penalty term to the loss function during training in order to improve generalization) is a widely diffused method. \emph{Adversarial training} \cite{szegedy2014} was the first defense to be proposed. It consists of enlarging the original dataset with a set of AEs, which is used to retrain the network. It has been criticized because it just shifts the problem to find new AEs.
Among others, it is worth citing \emph{defensive distillation} \cite{papernot2015distillation}, which trains a second, simpler network over soft targets (i.e., the output values of the original network) and \emph{deep contractive networks} \cite{gu2014deep}, which improve the defense performance of denoising convolutional autoencoders.

\paragraph{Auxiliary Tools} these approaches make use of an external tool to defend or detect AEs. Among these, Defense-GAN \cite{samangouei2018defensegan} and MagNet \cite{meng2017magnet} present good performance.

Many other works are present in the literature. The interested reader may refer to the reviews presented by Yuan et al. \cite{YuanTransactions}, and by Xu et al. \cite{AdversReview}.

\subsection{This work}
Although many of the defenses listed above show good performance, the work presented in this paper focuses on data modification techniques, since they are simple and computationally cheap to detect AEs, without changing or retraining the neural network. The proposed detection system can also be seen as an external tool aimed at detecting AEs, but without considering complex models (such as generative adversarial networks or additional neural networks) as previous works do.

Previous work used input transformations \cite{xie2017mitigating}, \cite{visionguard}, \cite{prakash}, \cite{guo2017countering} but, to the best of our records, none of them presented an extensive experimental evaluation to determine which transformations are the most effective in terms of detection rate. Furthermore, still to our records, no counter-measure for robust AEs has been presented before. This latter aspect is crucial, since any defense based on differentiable input transformations (as in the work of Xie et al. \cite{xie2017mitigating}) can be completely fooled by robust AEs.

This paper presents a novel method to detect robust AEs with input transformations, which, to our records, was never proposed in the literature. An experimental comparison with similar state-of-the-art methods is performed in Section \ref{s:exp}.


\section{Detection Algorithms} \label{s:proposed}
The objective of this work is to detect different kinds of AEs and evaluate the detection performance of four different architectures.
The AEs considered in this work can be classified 
into \textbf{(i)} \emph{standard}, i.e., those generated by the attacks reviewed at the beginning of Section~\ref{s:bkg}), and \textbf{(ii)} \emph{robust} ones, i.e., those that cannot be detected by input transformations.
Orthogonally, they can be also classified as \textbf{(i)} \emph{single-network}, i.e., those generated to attack just a single CNN, and \textbf{(ii)} \emph{multi-network}, i.e., those that are capable of attacking multiple CNNs subject to majority voting.
Note that this taxonomy allows defining four classes of AEs.

The \BASELINE and \VOTINGBASELINE architectures (Figures~\ref{f:single_net_detection} and~\ref{f:arch1}) are suited for the detection of standard AEs only, as they fail in detecting robust ones.
The former is suited for single-network AEs, while the latter for multi-network ones.

Other two architectures are proposed to detect both standard and robust AEs by leveraging a defense perturbation that is assumed not be known by the attacker. The \ENHANCED architecture is designed to detect single-network AEs, while the \VOTINGENHANCED architecture is designed to detect multi-network AEs by combining voting and defense perturbations.

%

\subsection{Baseline detection architectures}
The \BASELINE architecture has been considered to evaluate the detection performance of a set of input transformations. The input image $x \in \mathcal{X}$ (that might be an AE) is transformed by means of the chosen input transformation, producing $x'$. Then both images are fed into the network and a distance between the two output probability distributions is computed using the KL divergence. Since the KL divergence is not symmetric, it is not a proper distance. Hence, the actual distance is computed as the maximum KL of the two possible combinations:
\begin{equation} \label{e:KL}
D(f(x), f(x')) = \max \left \{\KL(f(x), f(x')), \KL(f(x'), f(x)) \right \}.
\end{equation}

The computed distance is then thresholded in order to classify the image as adversarial or not. Note that this is different from the approach proposed by Kantaros et al.~\cite{visionguard}, where the minimum of the two KL divergences is used. This is because all our experiments showed that the detection algorithm obtains a significantly better classification accuracy using the maximum, rather than the minimum. This can be explained by considering that the maximum is a more conservative estimate of the distance and has more impact on AEs, as they are characterized by more asymmetrical KL divergences than regular images.

The \VOTINGBASELINE architecture is also considered to assess the performance of a multi-network ensemble in detecting AEs. In particular, three pre-trained state-of-the-art CNNs are used, as shown in Figure \ref{f:arch1}. The three networks are subject to majority voting, which is a common algorithm for fault detection and exclusion and in general for fault-tolerant systems. Here, it is used to decide whether an input image is adversarial for the majority of the networks, meaning that as long as there are two networks detecting potential AEs, the voting algorithm classifies the input as adversarial.
The detection performance of such an architecture is discussed in Section \ref{s:exp}.

These two architectures are based on image transformations only and both fail in detecting robust AEs. The following section introduces new architectures to overcome this limitation. 

\subsection{Enhanced detection architectures: counter-measures against robust adversarial examples} \label{s:enhanced-archs}
AEs have great fooling power. During the experimental campaign conducted for this work, it was possible to find AEs that were robust to any kind of differentiable input transformation and even robust to any possible combination of four consecutive transformations, chosen between three different transformations. Also, randomization does not help: during the experiments it was possible to find AEs that are robust to additive Gaussian noise.

Since AEs have such great fooling power, the key idea of this work is to use the same crafting procedure to generate a \emph{defense perturbation}, which is basically a mask of pixels added to the input image at run-time. This perturbation is optimized to make robust AEs sensitive again to input transformations.

\begin{figure}[!t]
\centering
\includegraphics[width=\FIGSIZEXL]{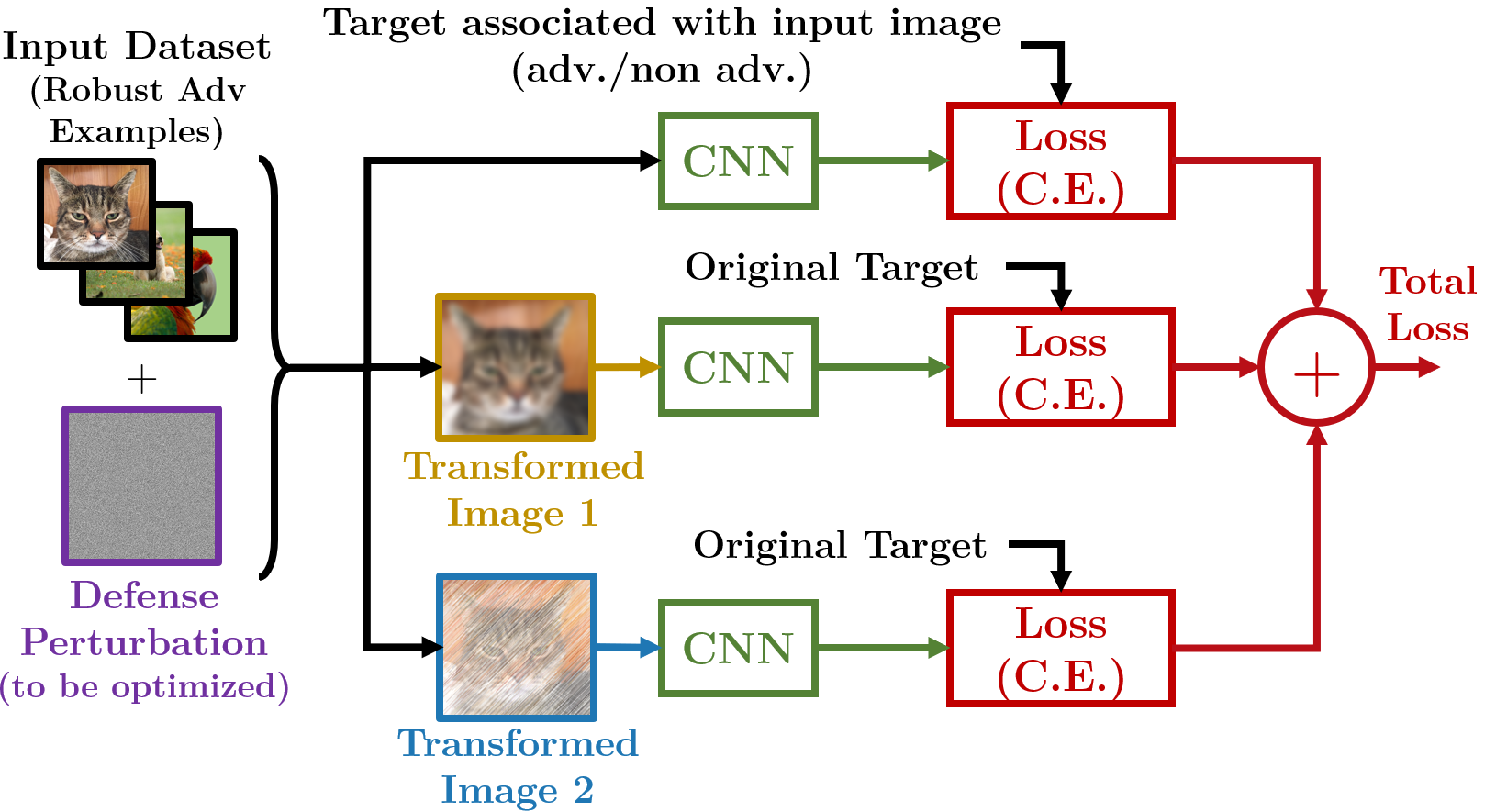}
\caption{\DEFENSEGEN architecture used to craft the defense perturbation.}
\label{f:defense_perturbation}
\end{figure}

The idea behind the generation of the defense perturbation is inspired by the one used to compute universal adversarial perturbations \cite{universal}.  This generation process is aimed at producing an image-agnostic pixel mask that is capable of removing the robustness to specific transformations, once it is added to a robust AE.
The defense perturbation is generated by using the architecture shown in Figure~\ref{f:defense_perturbation}, which is named \DEFENSEGEN, and takes as input a dataset constructed as follows. Given a set of $L$ original images and a set of $N$ input transformations $\{g_j(x, \theta_j), j=1, \ldots, N \}$, a robust AE is generated for each of the $L$ images and for each of the $N$ input transformations. All images, i.e., both the original and the adversarial ones, are then added to a dataset.
For each of the images $\tilde{x}$ in such a dataset,
the following optimization procedure is performed for $k_{max}$ steps:
\begin{equation} \label{e:defense-single-net}
\min_d [ \mathcal{L}(f(\tilde{x}+d), \tilde{t}) + \sum\limits_{j=1}^N\mathcal{L}(f(g_j(\tilde{x}+d; \theta_j)), t) + k\|d\|_2^2],
\end{equation}
where $t$ is the original (non-adversarial) target and $\tilde{t}$ is the target associated with image $\tilde{x}$. If $\tilde{x}$ is an AE, $\tilde{t}$ is the adversarial target, whereas if $\tilde{x}$ is a regular image, $\tilde{t}$ is the original target.
This iterative procedure outputs a mask $d$ that, when added to a robust AE $\tilde{x}$, produces a new image that is no longer robust to input transformations, and therefore can be used to detect AEs, just as in the standard case.

This property comes from the fact that the generation process of the defense perturbation follows the same logic for generating robust AEs, but reversed. The \ROBADVGEN (Equation \eqref{e:min_robust}) optimizes a perturbation, added to the original image, that pushes the prediction of the network towards the same label for both the non-transformed and the transformed images. Conversely, the \DEFENSEGEN (Equation \eqref{e:defense-single-net}) does the opposite: it pushes the prediction of the transformed image towards the original label (as it actually is with standard AEs), and the non-transformed image towards the corresponding label. Since the input images are both AEs and non-AEs, the corresponding label would be the adversarial label for AEs, and the original label for non-AEs. In this way, the defense perturbation learns the ``opposite average robust perturbation" that, added to a robust AE, removes its robustness.

If the optimization described in Equation~\eqref{e:min_robust} is used to generate an AE for each network and for each transformation, then the minimization of Equation~\eqref{e:defense-single-net} is aimed at finding the perturbation that is able to do the opposite, that is inhibiting the patterns that make the AEs robust to input transformations.

The downside of this method is that the generation of a dataset of robust AEs and the training of a defense perturbation are computationally expensive.
Another problem of this method is that it is suited only for robust AEs, while it has been experimentally found that has the property of making standard AEs robust. This happens because the role of the defense perturbation is to counteract the adversarial perturbation that makes an AE robust to input transformations.
When adding the defense perturbation to a standard AE, it introduces a perturbation that transforms standard AEs into robust ones, hence having a negative effect for the detection architecture.

\begin{figure}[!t]
\centering
\includegraphics[width=\FIGSIZEXL]{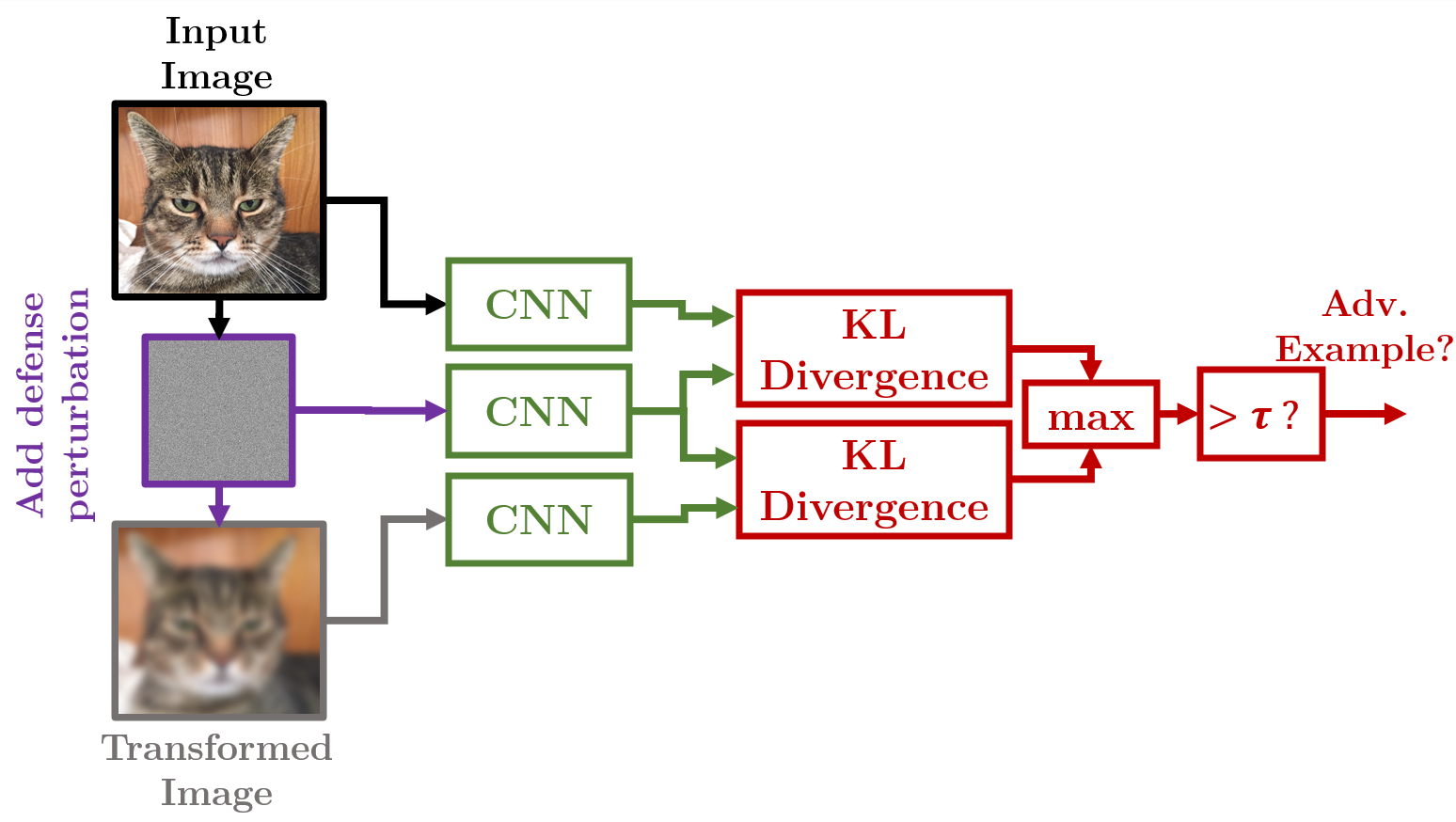}
\caption{\ENHANCED architecture for detecting robust adversarial examples.}
\label{f:arch2}
\end{figure}

Nevertheless, standard AEs are sensitive to any kind of perturbation, meaning that just the application of the mask is sufficient to discriminate AEs.
Given these observations, the detection of both standard and robust AEs can be performed using the \ENHANCED architecture illustrated in Figure~\ref{f:arch2}, where the mask is used as a first transformation (to detect standard AEs and to make robust AEs sensitive to input transformations), and a second transformation is used to detect robust AEs. Each of the two KL divergences is computed by Equation \eqref{e:KL}, taking the maximum between the two results.
With this counter-measure, robust AEs can be detected with simple input transformations at run-time and three inference operations with the same network.

Note that, since the defense perturbation $d$ is a differentiable transformation, an attacker could generate AEs that are robust to that specific mask. Being the mask not known in advance, the attacker might follow the same procedure described in this section to generate another defense mask, which has the same defensive properties, and then use it to craft AEs that are robust to that mask.
However, as one may expect, experiments show that different datasets used in input to the above procedure
lead to different defense perturbation masks.
Hence, the attacker should also know the exact data distribution used to generate the defense perturbation, which can easily be kept secret.

\begin{figure}[!t]
\centering
\includegraphics[width=\FIGSIZEXL]{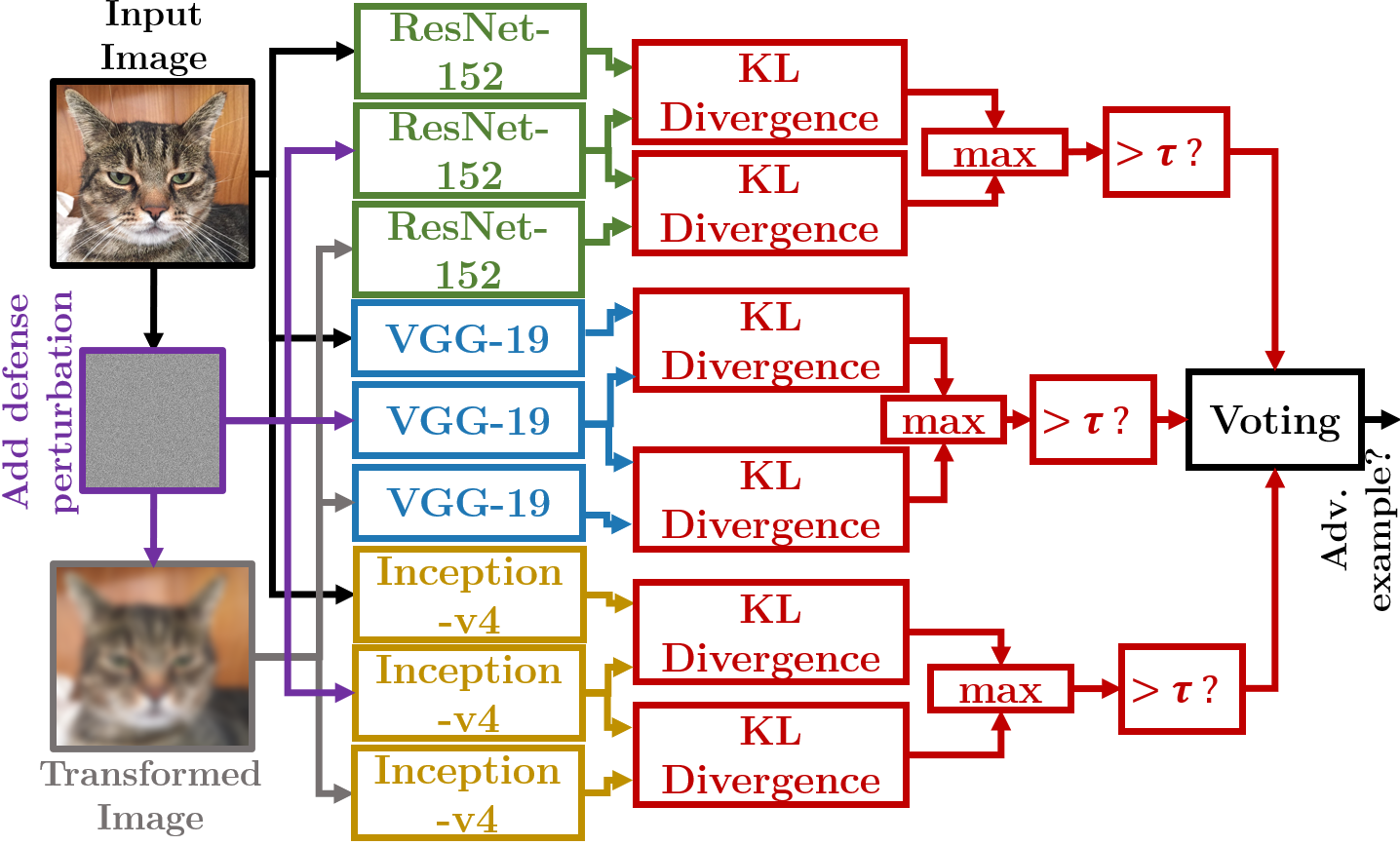}
\caption{The \VOTINGENHANCED detection architecture for detecting robust adversarial examples.}
\label{f:voting_enhanced}
\end{figure}

A further evolution of the detection architecture is finally proposed by combining the \ENHANCED architecture with majority voting, hence obtaining the architecture illustrated in Figure~\ref{f:voting_enhanced}, which is named \VOTINGENHANCED.
Under this latter architecture with voting over $M$ networks, the defense perturbation is generated as for the \ENHANCED architecture but optimizing
\begin{equation}
\min_d [\sum\limits_{i=1}^M \mathcal{L}(f_i(\tilde{x}+d), \tilde{t}) + \sum\limits_{j=1}^N\sum\limits_{i=1}^M\mathcal{L}(f_i(g_j(\tilde{x}+d; \theta_j)), t) + k\|d\|_2^2].
\end{equation}

The defense perturbation generated in this way results to be effective for multi-network robust AEs.

\section{Experimental Results} \label{s:exp}
This section presents the results obtained from an extensive experimental evaluation carried out to test the performance of the approaches proposed in this paper to detect AEs. After describing the experimental setting, the input transformations used for the evaluations are introduced, together with each characteristic parameter and their range. Then, the effect of each transformation on the accuracy of the networks is reported. The performance of the different detection methods is presented by first considering the \BASELINE and \VOTINGBASELINE architectures, and then the \ENHANCED and \VOTINGENHANCED architectures, which include the defense perturbation.

\subsection{Experimental setting}
Three CNNs were selected for the experiments: \textbf{(i)} VGG-19~\cite{vgg}; \textbf{(ii)} Resnet-v2-152~\cite{resnet}; and \textbf{(iii)} Inception-v4~\cite{inceptionv4}.
These networks were chosen as they are among the most used, top-performing CNNs for visual recognition. They were pre-trained on the ImageNet dataset and downloaded from the tf-slim library~\cite{TFSlim}.

All the experiments presented in this paper were performed on a Nvidia DGX station, 
composed of 4 Tesla-v100 GPUs, with 32GB of RAM each.
The code for the experiments was written in Python 3 using Tensorflow 1.15 (configured to use GPUs). The ImageNet dataset was downloaded from Kaggle.
Under this setting, the generation of a single multi-net robust AE took 250 seconds.

\subsection{Input Transformations} \label{s:transformations}
The image transformations considered in this work are several and can be divided into three different groups. Each transformation comes with a parameter that varies in a certain range.
\paragraph{Topological transformations} They include the basic affine transformations that can be expressed in the form of a warping matrix $T \in {\rm I\!R}^{3\times 3}$. Each of such transformations can be defined as $\nu' = T \nu$, where $\nu = [u, v, 1]^T$ and $\nu' = [u', v', 1]^T$ represent the homogeneous coordinates of a pixel of the original and the transformed images, respectively.
The transformations of this type considered in this work are:
\begin{itemize}
  \item \textbf{Translation}: horizontal and vertical translation parameters are combined into a single diagonal translation to simplify the parametric search. Range: $[-45$ px$, +45$ px$]$;
  \item \textbf{Rotation}. Range: $[-25^{\circ}, 25^{\circ}]$;
  \item \textbf{Horizontal Shear}: it is expressed by the transformation \footnotesize$T=\begin{bmatrix}
  1 & s_x & 0 \\ 0 & 1 & 0 \\ 0 & 0 & 1
\end{bmatrix}$, \normalsize where $s_x$ is the shear parameter with range $[-0.175, 0.175]$;
  \item \textbf{Scale}: it is expressed by the transformation $T=\text{diag}([s, s, 1])$, where $s$ is the scale parameter with range $[0.875, 1.175]$;
  \item \textbf{Mirror} (no parameter).
\end{itemize}

\paragraph{Appearance Transformations} They include those transformations that change the appearance of the image, with no topological changes. The ones considered in this work are:
\begin{itemize}
  \item \textbf{Average Blur}. Range: $[2\times 2$ kernel$, 6\times 6$ kernel$]$);
  \item \textbf{Brightness Change}: it adds the value of the parameter to the intensity of the image, pixel-wise. Range: $[-35, +35]$;
  \item \textbf{Contrast Change}: it scales the pixel-wise intensity of the image by the parameter value. Range: $[0.875, 1.125]$.
\end{itemize}

\paragraph{Special} They include two uncategorized transformations:
\begin{itemize}
  \item \textbf{Bit Depth Reduction}: it changes the number of bits used to represent intensity of the pixels. Range: [4 bits, 7 bits];
  \item \textbf{Gaussian Noise}: it adds a Gaussian noise to each pixel of the input image (no parameters\footnote{The additive Gaussian noise is with 0 mean and standard deviation set equal to the adversarial strength $\epsilon=0.05$ of the AEs, to obtain a fair comparison between attacker and defender capabilities.}).
\end{itemize}

It is worth noting that all the transformations used are differentiable with respect to the input image. This is crucial, since this work assumes white-box robust AEs. Other more complex non-differentiable transformations were used in the literature, but they are out of the scope of this work, since a black-box approach should be used to craft AEs that are robust to those transformations.

\subsection{Performance Metrics}
AEs detection is a binary classification problem. A classical way to evaluate the performance of balanced binary classifiers with a decision threshold is through the Receiver Operating Characteristic (ROC), which expresses, for different values of the threshold, the fraction of \emph{true positives} (TP), i.e., AEs detected as AEs, against the fraction of \emph{false positives} (FP), i.e., non-AEs detected as AEs.

The baseline performance, i.e., the performance of a random classifier that detects as many TPs as FPs for any value of the threshold, is represented in a ROC plot by the linear pattern with slope 1 (passing through $(0, 0)$ and $(1, 1)$), while the best performance is reached when the classifier achieves TP rate 1 and FP rate 0. A more compact way to represent the performance is through the Area Under Curve (AUC) of a ROC graph: perfect classification performance has AUC of 1, while baseline performance has AUC equal to 0.5.

When using this performance indicator, the information about exact TP/FP ratio is lost. However, this is not important when dealing with detection systems with AUC close to 1, since they will produce very similar-looking ROC graphs, all passing close to the point (1, 0). Conversely, when the classification performance is poor, the ROC curve has much more ``freedom", and could assume several different shapes with the same AUC result. Since the focus of this work is to achieve a high-performance AEs detection, there is no interest in studying the actual ROC curve of poorly performing detection systems. Solutions with AUC $\geq$ 0.95 can be considered satisfying for our purposes.

Results are presented for multiple transformations. In all the following plots, the performance value (accuracy or AUC of ROC) is on the y-axis, while the x-axis refers to the normalized value of the parameter used to control the transformations.
Note that some transformations are controlled by a parameter that varies in a symmetric range (e.g., see the case of translation), while others are not. The ranges of the former are of the form $[\gamma-\alpha, \gamma+\alpha]$ and each parameter $\theta \in [\gamma-\alpha, \gamma+\alpha]$ is reported on the x-axis as $(\theta-\gamma)/\alpha$, i.e., obtaining a normalized representation that varies from -1 to +1. Conversely, non-symmetric ranges are of the form $[\alpha, \beta]$, with $\alpha<\beta$, and each parameter $\theta \in [\alpha, \beta]$ is reported
on the x-axis as $\theta/\beta$, hence obtaining a normalized value that varies from $\alpha/\beta$ to 1 only.
%
%
Mirror and Gaussian noise are parameter-free transformations and their performance is represented by a single point with value 0 on the x-axis.

\subsection{Accuracy Drop}
\begin{figure}[!t]
\centering
\includegraphics[width=\FIGSIZE]{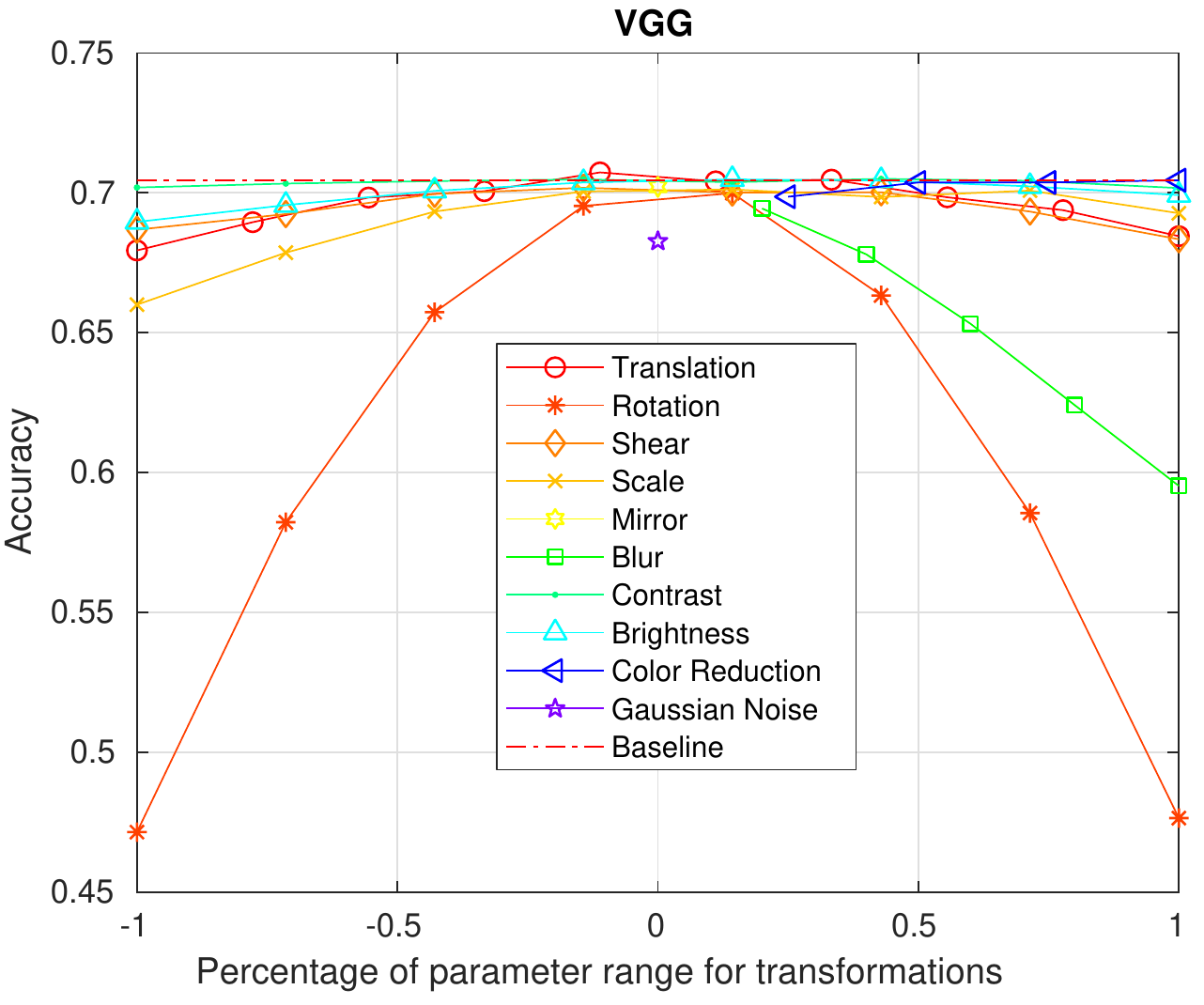}
\caption{Accuracy of VGG-19 when using input transformations.}
\label{f:accuracy}
\end{figure}

A major drawback of using input image transformations for AEs detection is that they cause an accuracy drop for the classifier.
The baseline accuracy is evaluated without input transformation on a validation subset of ImageNet composed of 10 images for each class, for a total of 10k images. For each transformation, the accuracy is computed for discretized values within the ranges of each parameter and reported in Figure~\ref{f:accuracy} for VGG-19. The results obtained with the other networks show a very similar pattern for all the transformations, hence they are not reported for space limitations.

\subsection{Adversarial Examples Datasets}
Several datasets of AEs have been produced to test the performance of the proposed approaches and generate the defense perturbations. Each dataset consists of multi-network AEs.
For standard AEs, two datasets have been generated, one with the L-BFGS attack and one for the CW attack, each containing 1000 samples, one for each class of the ImageNet dataset. For robust AEs, four different datasets have been generated, each serving a different purpose:
\begin{itemize}
\item $M_{\text{test}}$. It is generated to test the performance of the \ENHANCED and \VOTINGENHANCED architectures. It includes adversarial samples generated starting from nine samples of the first 100 classes and two for the other 900 classes, for a total of 2700 AEs. This dataset is much larger than the others because it is crucial to evaluate whether the proposed counter-measure is effective in general, meaning that it has to generalize both for a wide distribution of AEs belonging to the same class and for each class.
\item $M_{\text{def-gen}}$. It is the one used to generate the defense perturbation, referred to as $D_{\text{def}}$ in the following, used in the \ENHANCED and \VOTINGENHANCED architectures, as detailed in Section~\ref{s:enhanced-archs}.
\item $M_{\text{att-gen}}$. This dataset is used to simulate the case in which an attacker tries to attack the \ENHANCED and \VOTINGENHANCED architectures. In this case, since the attacker is assumed not to know $D_{\text{def}}$, he/she has to first generate its own defense perturbation (referred to as $D_{\text{att}}$) before attempting at generating AEs that are robust to the enhanced detection architectures. The $M_{\text{att-gen}}$ dataset serves the purpose of generating $D_{\text{att}}$ with the approach detailed in Section~\ref{s:enhanced-archs}.
\item $M_{\text{att}}$. It is the dataset generated to be robust to both input transformations and defense perturbation $D_{\text{att}}$. It is used to experimentally confirm that it is not possible to generate AEs that are robust to the \ENHANCED and \VOTINGENHANCED architectures, provided that the attacker does not know the exact AEs used to generate the defense perturbation $D_{\text{def}}$.

\end{itemize}

All the multi-network robust AEs are generated by considering (i) one transformation for each class discussed in Section~\ref{s:transformations}, namely translation, blur, and Gaussian noise;
and (ii) the cross-entropy loss function (as for the L-BFGS attack).

These three transformations proved to be sufficient to make the AE robust to the entire set of considered transformations, hence making possible to generate robust AEs with a limited computational effort. 


%
%

All the generated datasets are summarized in Table \ref{t:datasets}. The actual size of the dataset is twice the one showed in the table, since it also includes the original images.

Except for the L-BFGS dataset, which was optimized over 250 epochs, all the other datasets were optimized over 500 epochs. For every dataset the Adam optimizer was used, with learning rate of 0.1, and a fixed adversarial strength $\epsilon = 0.05$ was chosen.

\begin{table}[!t]
\renewcommand{\arraystretch}{1.3}
\setlength{\tabcolsep}{3pt}
\caption{Adversarial Examples datasets generated. DP1 refers to the Defense Perturbation used for detection, whereas DP2 refers to the defense perturbation generated to craft defense-robust adversarial examples.}
\label{t:datasets}
\centering
\begin{tabular}{|c||c|c|c|}
\hline
\textbf{Dataset} & \textbf{Robust to...} & \textbf{Samples} & Notes \\
\hline
\hline
\begin{tabular}{@{}c@{}}Standard \\(L-BFGS) \end{tabular} & - & 1000 & 1 per class\\
\hline
\begin{tabular}{@{}c@{}}Standard CW \\(CW) \end{tabular} & - & 1000 & 1 per class\\
\hline
\begin{tabular}{@{}c@{}}Multi-robust \\($M_{\text{test}}$) \end{tabular} & \begin{tabular}{@{}c@{}} Translation \\ Blur \\ Gaussian Noise \end{tabular}  & 2700 & \begin{tabular}{@{}c@{}}9 for the first 100 classes \\ 2 for the others \\Used for tests\end{tabular}\\
\hline
\begin{tabular}{@{}c@{}}Multi-robust \\($M_{\text{def-gen}}$) \end{tabular} & \begin{tabular}{@{}c@{}} Translation \\ Blur \\ Gaussian Noise \end{tabular}  & 1000 & \begin{tabular}{@{}c@{}}1 per class \\Used for $D_{\text{def}}$ generation\end{tabular}\\
\hline
\begin{tabular}{@{}c@{}}Multi-robust \\($M_{\text{att-gen}}$) \end{tabular} & \begin{tabular}{@{}c@{}} Translation \\ Blur \\ Gaussian Noise \end{tabular}  & 1000 & \begin{tabular}{@{}c@{}}1 per class \\Used for $D_{\text{att}}$ generation \\(to craft dataset $M_{\text{att}}$)\end{tabular}\\
\hline
\begin{tabular}{@{}c@{}}Defense-robust \\($M_{\text{att}}$) \end{tabular} & \begin{tabular}{@{}c@{}} Translation \\ Blur \\ Gaussian Noise \\ $D_{\text{att}}$ \end{tabular}  & 1000 & 1 per class \\
\hline
\end{tabular}
\end{table}

\subsection{Detection of adversarial examples with only input transformations}
\begin{figure}[!ht]
\centering
\includegraphics[width=\FIGSIZE]{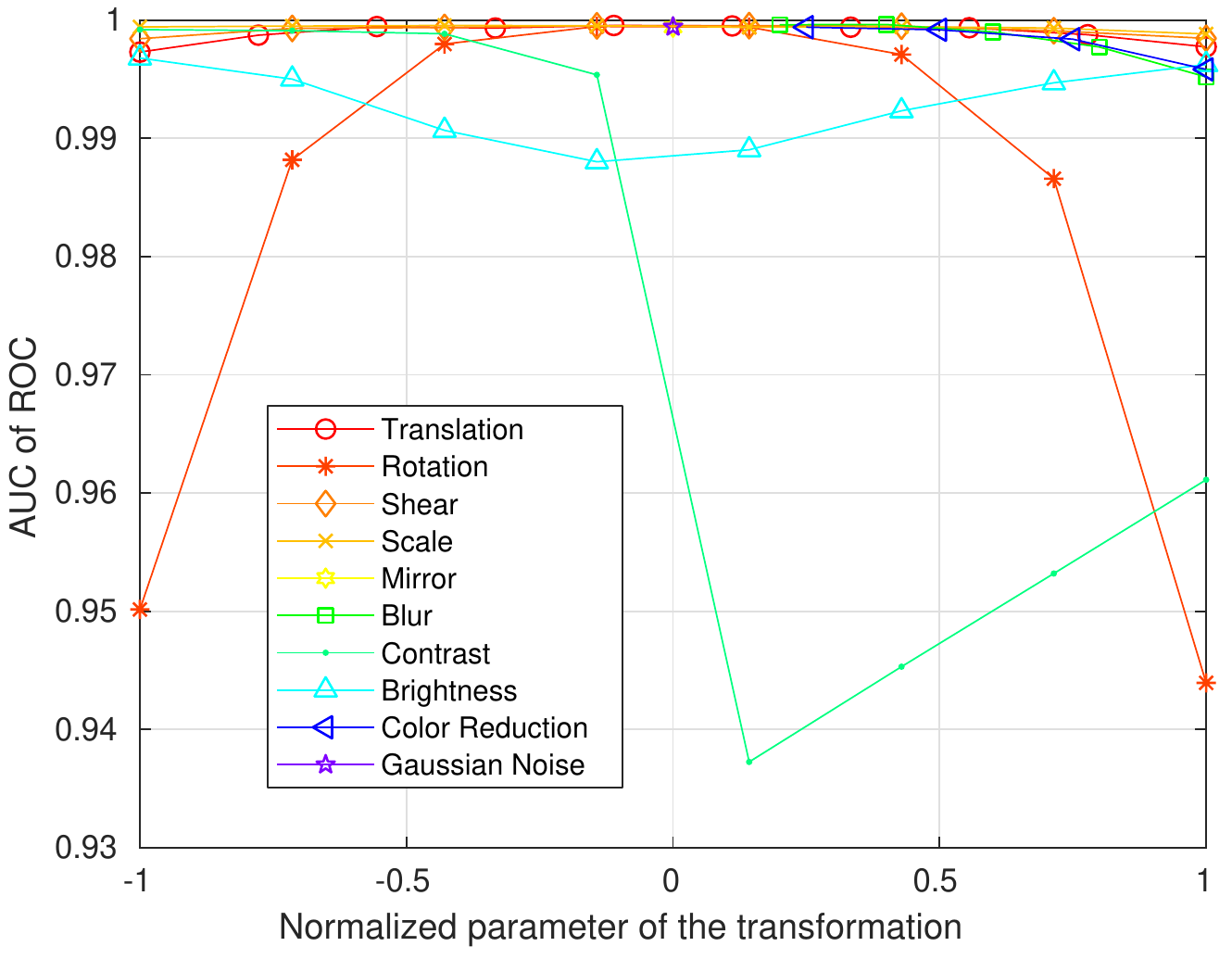}
\caption{Performance of the \VOTINGBASELINE architecture in detecting standard adversarial examples (L-BFGS) for each input transformation.}
\label{f:results_standard}
\end{figure}

\begin{figure}[!ht]
\centering
\includegraphics[width=\FIGSIZE]{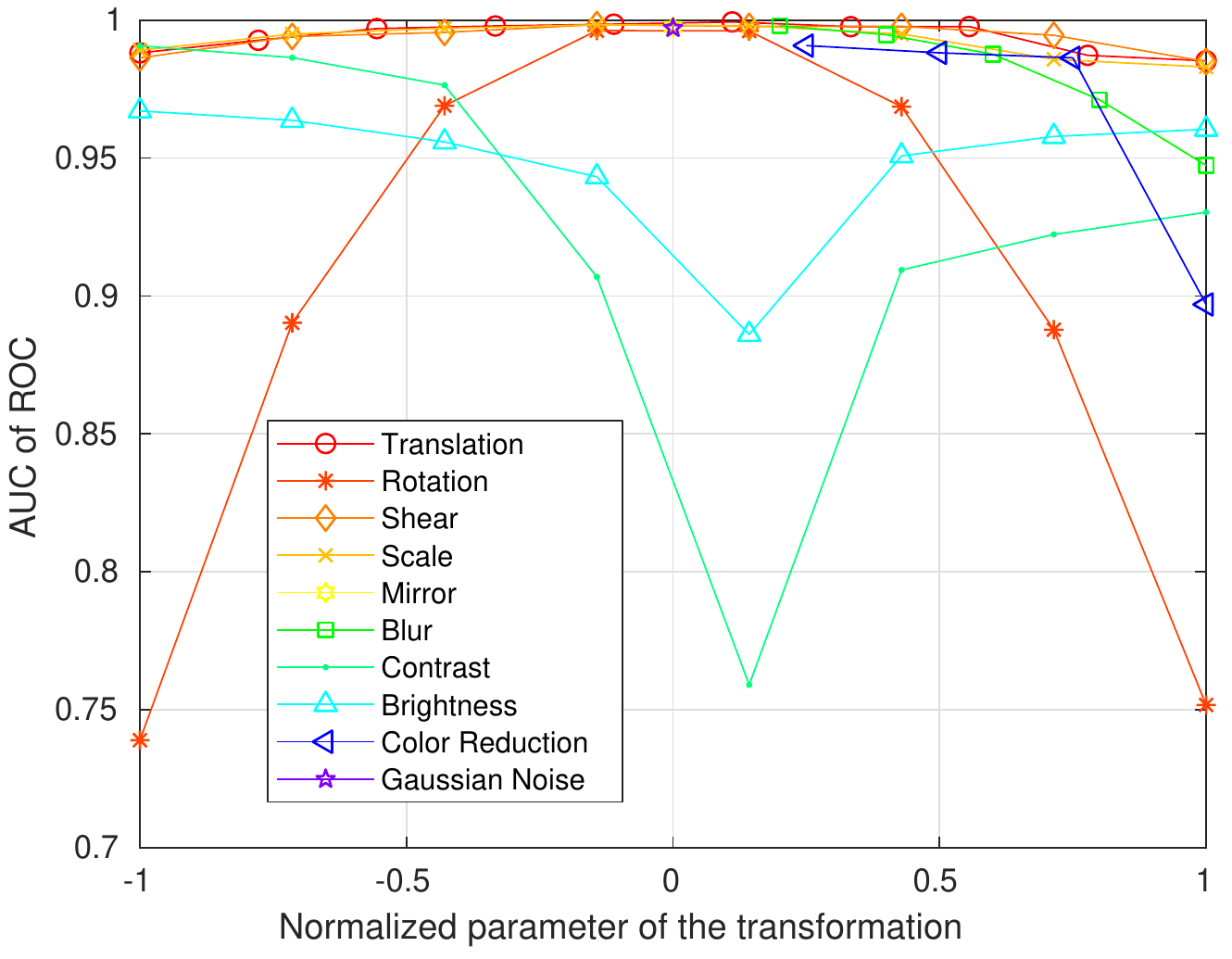}
\caption{Performance of the \VOTINGBASELINE architecture in detecting standard adversarial examples (CW) for each input transformation.}
\label{f:results_CW}
\end{figure}

This subsection discusses the performance of the \VOTINGBASELINE architecture (Figure~\ref{f:arch1}). When dealing with standard AEs, this architecture exhibits a very good performance, as it can be seen from Figures \ref{f:results_standard} (L-BFGS attack) and \ref{f:results_CW} (CW attack). The AEs generated with the CW attack resulted to be harder to detect and, for some transformation, the performance degrades quicker for higher parameter values. Clearly, some of the transformations are less suitable than others for this kind of detection. Contrast and brightness change, color reduction, and rotation show poorer performance with respect to other transformations (e.g., other topological transformations, blur and Gaussian noise).

\begin{figure}[!ht]
\centering
\includegraphics[width=\FIGSIZE]{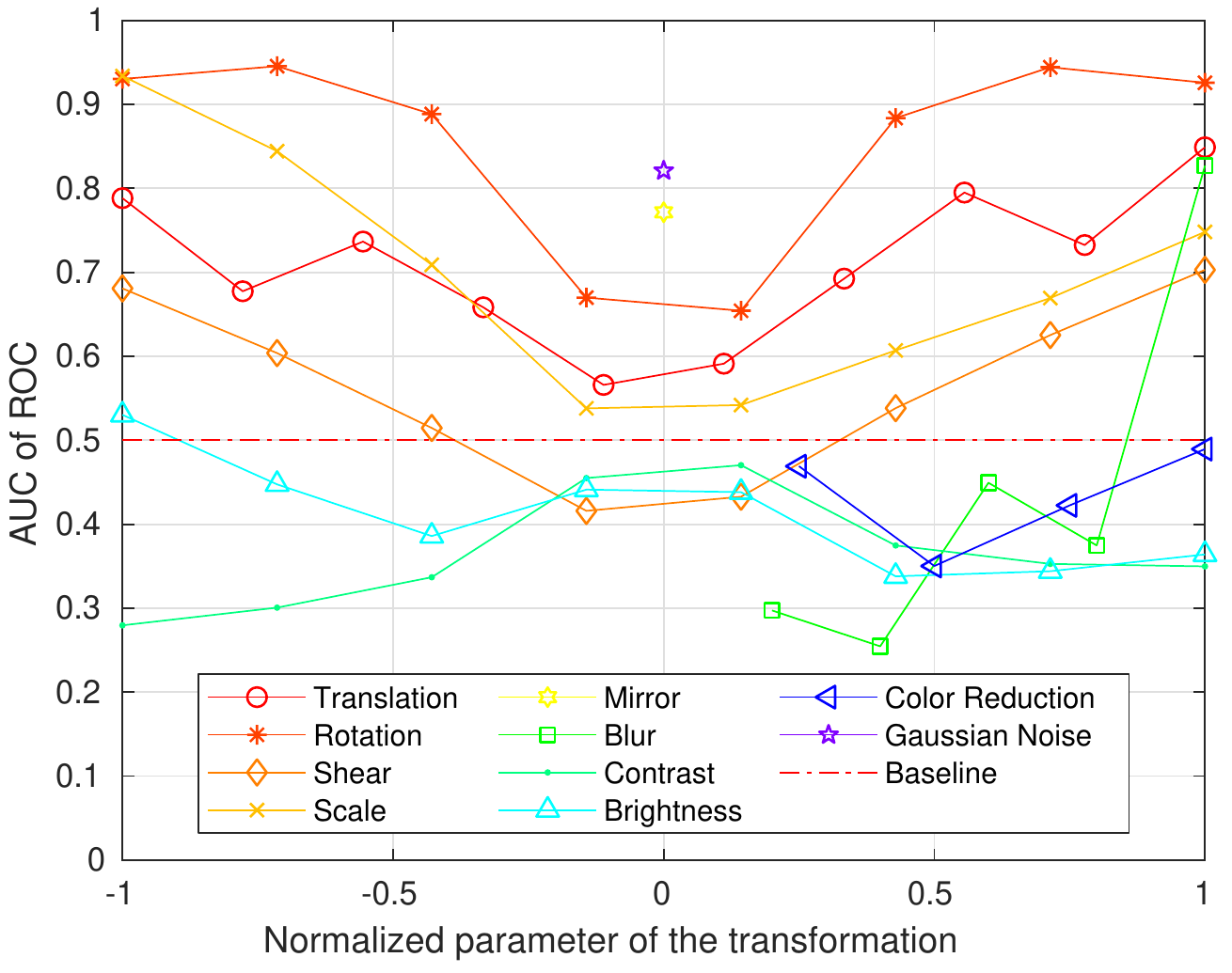}
\caption{Performance of the \VOTINGBASELINE architecture in detecting robust adversarial examples (translation, blur, Gaussian noise) for each input transformation.}
\label{f:results_robust}
\end{figure}

Very different results are obtained when presenting robust AEs to the \VOTINGBASELINE architecture. These results are summarized in Figure~\ref{f:results_robust}. As expected, the detection performance is much worse. What is interesting is that, even if the AEs were chosen to be robust to three transformations, the performance of all the other transformations drops, especially when dealing with small parametric values. This can be explained with the fact that transformations with small parameter values are similar to each other, especially when dealing with transformations of the same category (e.g., scale and shear have comparable detection performance with respect to translation).

Some transformations resulted to be even worse than a random detection algorithm, meaning that even a coin toss shows better AEs detection performance. The experiment shows that, when dealing with robust AEs, differentiable input transformations cannot be used in this naive detection system. Not even noise (Gaussian in this case) is a safe choice: even though it showed better performance with respect to other transformations, its AUC drops to about 0.82, with a ROC curve that saturates (i.e., 100\% of TP) for about 50\% of FP.

Further tests were conducted to explore the effect of cascaded transformations: the usage of a series of randomly-selected transformations can help when dealing with this kind of robust AEs, but it is also still possible to craft AEs that are robust to any possible combination of cascaded transformations. In our experiment, this has been verified via cascades of four randomly-selected transformations (picked from the three transformations used before, i.e., translation, blur and Gaussian noise). These results were omitted for space limitations.

\begin{figure}[!ht]
\centering
\includegraphics[width=\FIGSIZE]{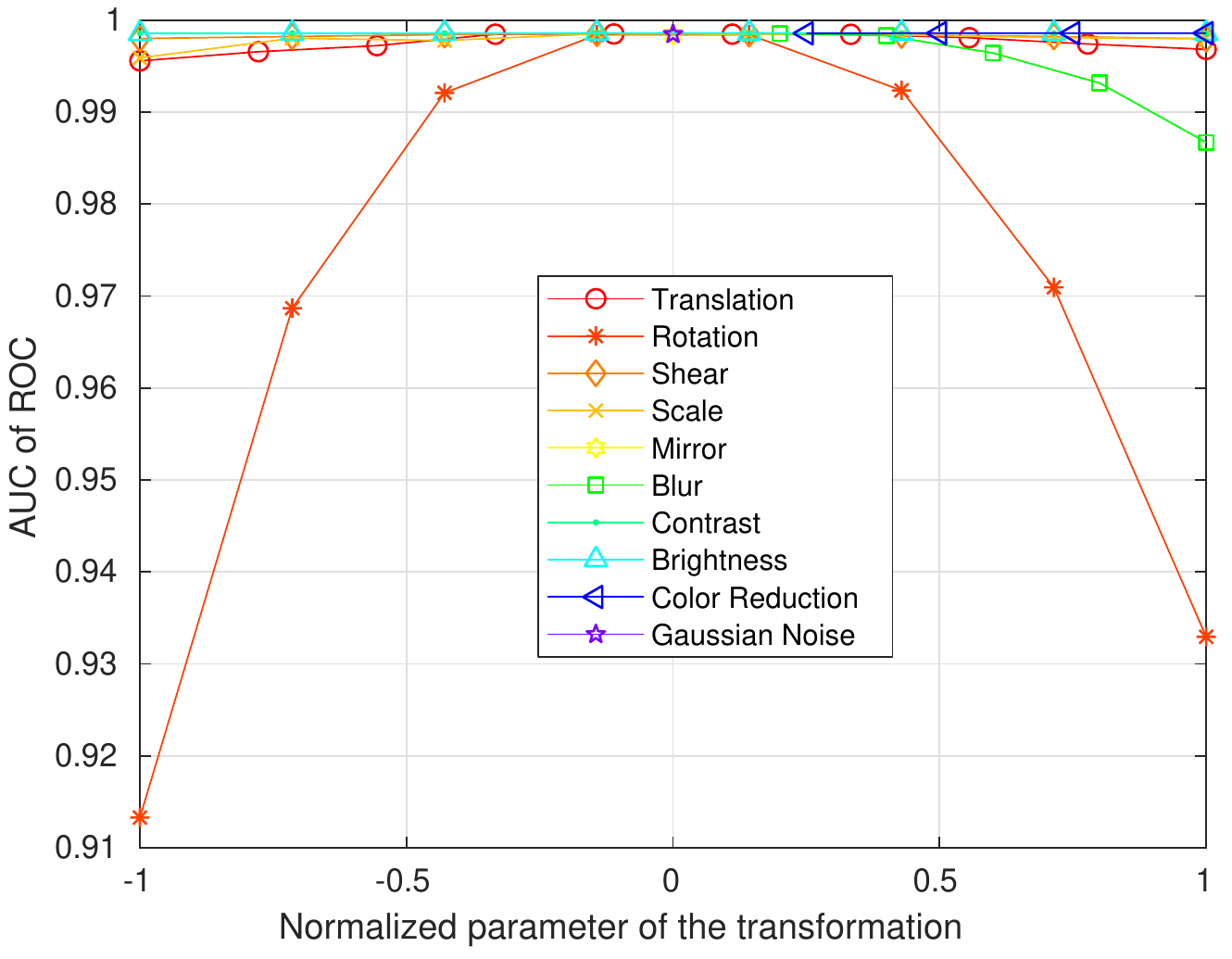}
\caption{Performance of the \VOTINGENHANCED architecture in detecting standard adversarial examples (L-BFGS) for each input transformation.}
\label{f:results_mask_standard}
\end{figure}

\begin{figure}[!ht]
\centering
\includegraphics[width=\FIGSIZE]{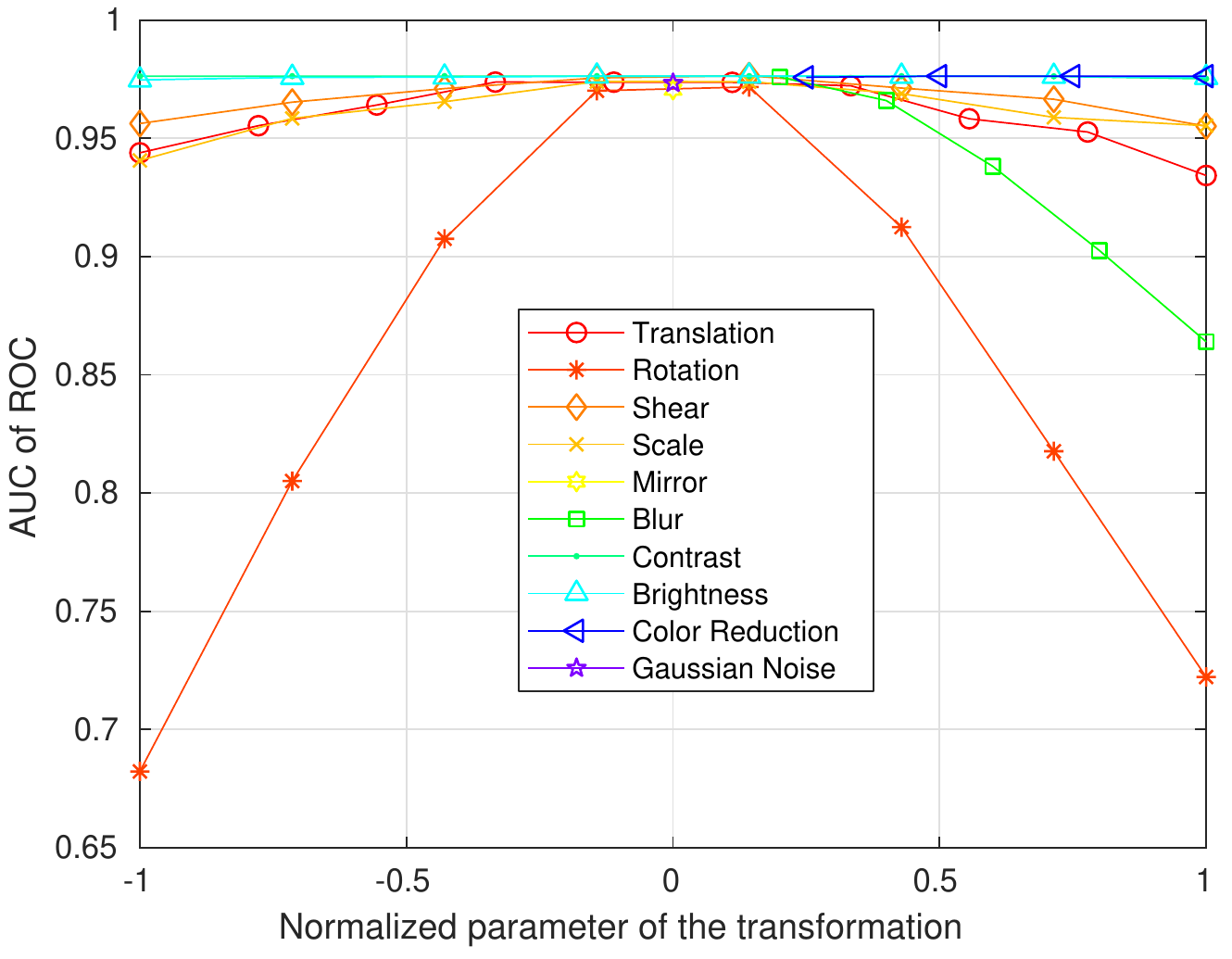}
\caption{Performance of the \VOTINGENHANCED architecture in detecting standard adversarial examples (CW) for each input transformation.}
\label{f:results_mask_CW}
\end{figure}

\begin{figure}[!ht]
\centering
\includegraphics[width=\FIGSIZE]{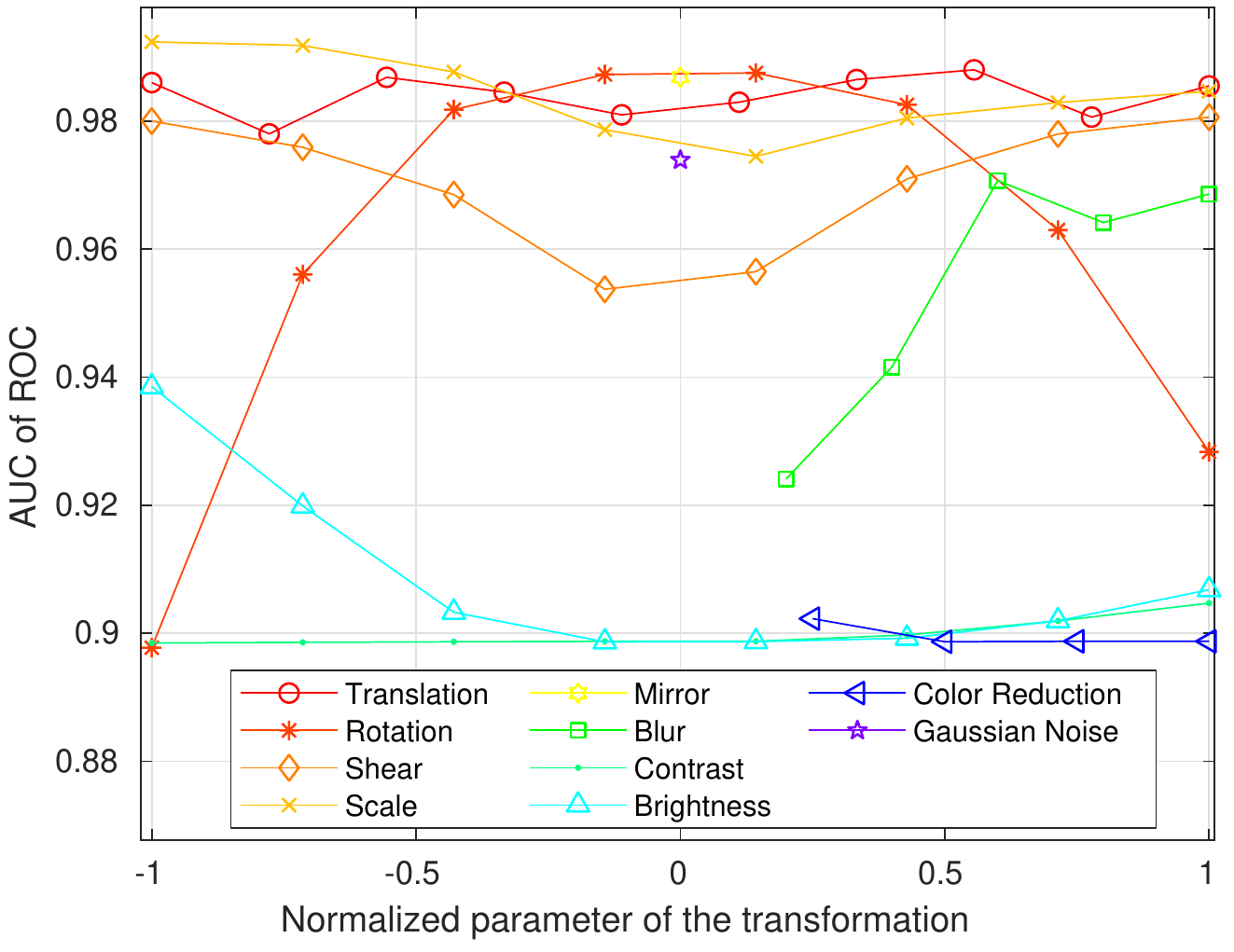}
\caption{Performance of the \VOTINGENHANCED architecture in detecting robust adversarial examples (set $M_{\text{test}}$) for each input transformation.}
\label{f:results_mask_robust}
\end{figure}
\subsection{Detection of adversarial examples using the defense perturbation}
The results of the previous subsection showed that an attacker unaware of the specific input transformation used for detection can still craft robust AEs that jeopardize the detection system. This subsection reports the results achieved by applying the defense perturbation with the \VOTINGENHANCED architecture illustrated in Figure \ref{f:voting_enhanced}.

As explained in Section~\ref{s:proposed}, the defense perturbation is effective not only for detecting standard AEs, but also for making robust AEs sensitive to input transformations again.
The AEs used to craft the defense perturbation are generated from images different than those used to test the performance of the detection system). The same three transformations were chosen (translation, blur, and Gaussian noise).



The performance of the \VOTINGENHANCED architecture in detecting standard AEs with a defense perturbation is shown in Figure \ref{f:results_mask_standard} for the L-BFGS attack, and in Figure \ref{f:results_mask_CW} for the CW attack, while the results for robust AEs are shown in Figure \ref{f:results_mask_robust}.
Interestingly, as observed for the results shown in Figure~\ref{f:results_robust}, although only three transformations were used to craft the defense perturbation, all the other transformations exhibited a performance boost from the use of the defense perturbation.

Another set of experiments has been performed to assess the robustness of the defense perturbation. As
mentioned in Section~\ref{s:enhanced-archs}, the attacker does not have access to the defense perturbation $D_{\text{def}}$ nor to the specific dataset $M_{\text{def-gen}}$ used to generate it. Anyway, he/she can still try to generate another defense perturbation $D_{\text{att}}$ from a different (but presumably similar) dataset $M_{\text{att-gen}}$ with the purpose of generating a set $M_{\text{att}}$ of AEs that are robust to $D_{\text{def}}$.

\begin{figure}[!t]
\centering
\includegraphics[width=\FIGSIZE]{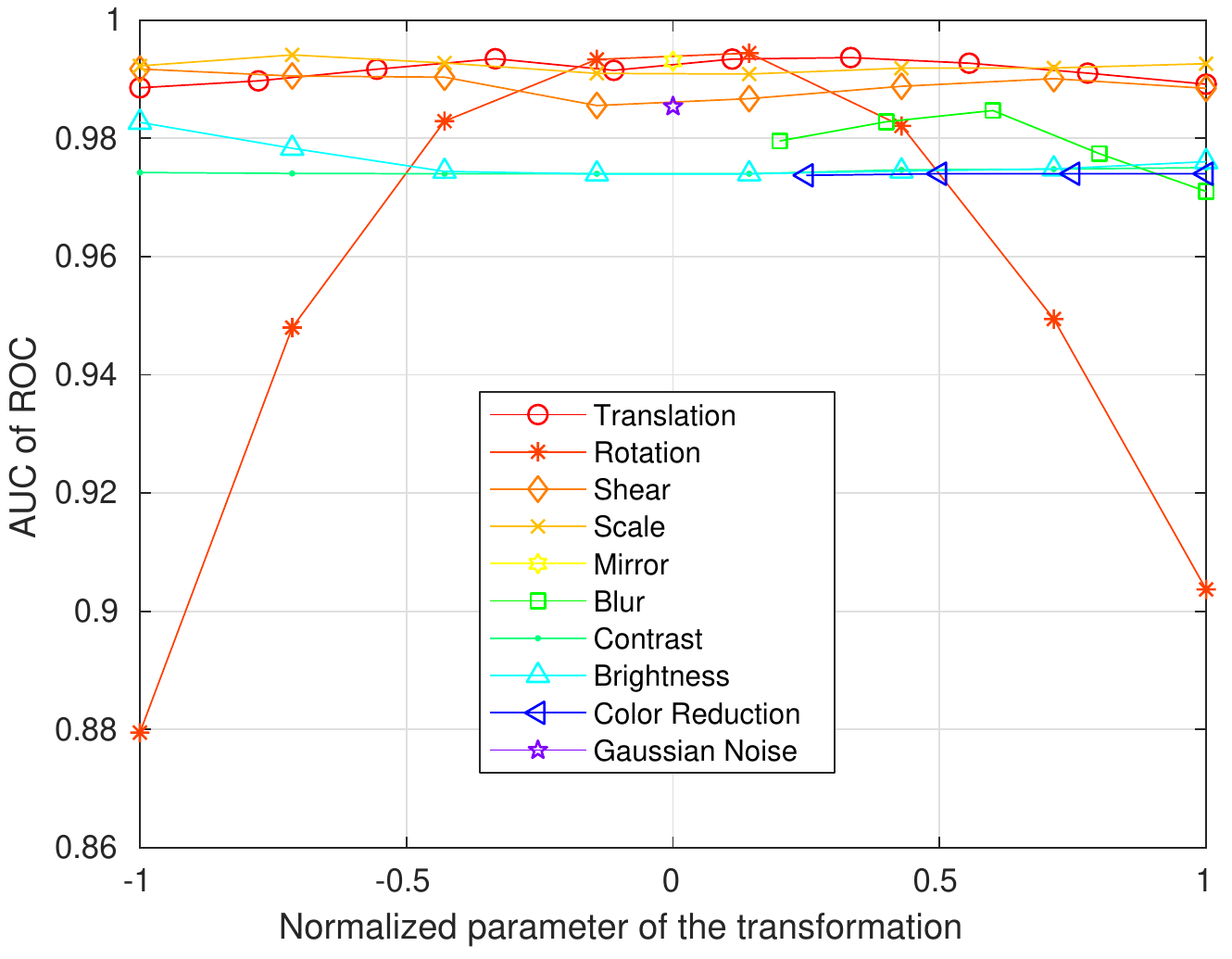}
\caption{Performance of the \VOTINGENHANCED architecture in detecting robust adversarial examples (set $M_{\text{att}}$) for each input transformation.}
\label{f:results_mask_maskrobust}
\end{figure}

The results of the detection of $M_{\text{att}}$ are reported in Figure \ref{f:results_mask_maskrobust}, which shows that
the AEs prepared by the attacker can still be detected with $D_{\text{def}}$ in the \VOTINGENHANCED architecture.
This makes the \VOTINGENHANCED architecture an effective
counter-measure for robust AEs, making them sensitive to input perturbations, while maintaining good detection performance on standard AEs.

\subsection{Effect of Voting}
This subsection evaluates the effect of voting used in the \VOTINGENHANCED architecture and shows how, in most cases, it increases the detection accuracy with respect to the \ENHANCED architecture, which considers a single network.

To present the results in a compact way, the effect of voting is reported as the minimum improvement of the AUC computed over the parametric range, for each transformation and for each network. The change is reported as a percentage of the total area (i.e., 1) for an easier reading. Positive values of the AUC change are to be considered as improvements of the \VOTINGENHANCED architecture with respect to the \ENHANCED architecture (i.e., with a single network). Conversely, negative values represent worse AUC.
The minimum change in performance over the parametric range is chosen because it indicates the worst-case improvement due to voting.


The results of this test are summarized in Table \ref{t:total_results}.
It is worth observing that voting helps in most of the cases, except for rotation, where the performance of the resulting voting system is heavily affected by the difference in performance between the networks.  
\begin{table}[!t]
\renewcommand{\arraystretch}{1.3}
\setlength{\tabcolsep}{4pt}
\caption{Minimum change of AUC of ROC (percentual of total AUC = 1) introduced by the \VOTINGENHANCED architecture with respect the \ENHANCED architecture. Results are shown for each transformation and for each dataset of adversarial examples; the rows in each cell represent the minimum change in AUC with respect to each network (VGG-19, Resnet-v2-152, and Inception-v4 respectively).}

\label{t:total_results}
\centering
\begin{tabular}{|c||c|c|c|c|}
\hline
 & L-BFGS & CW & $M_{\text{test}}$ & $M_{\text{att}}$\\
\hline
\hline
Translation & \begin{tabular}{@{}c@{}} +0.71 \\ +0.96 \\ +0.18 \end{tabular}
& \begin{tabular}{@{}c@{}} +4.90 \\ +2.74 \\ +0.50 \end{tabular}
& \begin{tabular}{@{}c@{}} +2.36 \\ +2.13 \\ +0.47 \end{tabular}
& \begin{tabular}{@{}c@{}} +2.05 \\ +2.37 \\ -0.01 \end{tabular}  \\
\hline
Rotation & \begin{tabular}{@{}c@{}} +0.78 \\ +1.07 \\ -3.53 \end{tabular}
& \begin{tabular}{@{}c@{}} +4.94 \\ -2.15 \\ -6.28 \end{tabular}
& \begin{tabular}{@{}c@{}} +3.41 \\ +2.23 \\ -2.92 \end{tabular}
& \begin{tabular}{@{}c@{}} +2.46 \\ +2.52 \\-4.88 \end{tabular}  \\
\hline
Shear & \begin{tabular}{@{}c@{}} +0.80 \\ +0.83 \\ +0.20 \end{tabular}
& \begin{tabular}{@{}c@{}} +5.13 \\ +2.59 \\ +0.78 \end{tabular}
& \begin{tabular}{@{}c@{}} +3.65 \\ +2.71 \\ +1.05 \end{tabular}
& \begin{tabular}{@{}c@{}} +2.94 \\ +2.37 \\ -0.04 \end{tabular}  \\
\hline
Scale & \begin{tabular}{@{}c@{}} +0.76 \\ +0.95 \\ +0.22 \end{tabular}
& \begin{tabular}{@{}c@{}} +4.95 \\ +2.90 \\ +0.41 \end{tabular}
& \begin{tabular}{@{}c@{}} +2.27 \\ +2.08 \\ +0.48 \end{tabular}
& \begin{tabular}{@{}c@{}} +2.37 \\ +2.25 \\ +0.12 \end{tabular}  \\
\hline
Mirror & \begin{tabular}{@{}c@{}} +0.75 \\ +1.07 \\ +0.19 \end{tabular}
& \begin{tabular}{@{}c@{}} +4.81 \\ +3.23 \\ +0.82 \end{tabular}
& \begin{tabular}{@{}c@{}} +2.28 \\ +1.74 \\ +2.03 \end{tabular}
& \begin{tabular}{@{}c@{}} +2.10 \\ +1.85 \\ +0.38 \end{tabular}  \\
\hline
Blur & \begin{tabular}{@{}c@{}} +0.85 \\ +0.88 \\ +0.26 \end{tabular}
& \begin{tabular}{@{}c@{}} +5.74 \\ +2.56 \\ +1.54 \end{tabular}
& \begin{tabular}{@{}c@{}} +3.23 \\ +5.03 \\ +0.18 \end{tabular}
& \begin{tabular}{@{}c@{}} +4.43 \\ +2.97 \\ -0.20 \end{tabular}  \\
\hline
Contrast & \begin{tabular}{@{}c@{}} +0.85 \\ +0.78 \\ +0.19 \end{tabular}
& \begin{tabular}{@{}c@{}} +5.78 \\ +2.23 \\ +1.30 \end{tabular}
& \begin{tabular}{@{}c@{}} +0.24 \\ +5.88 \\ +1.59 \end{tabular}
& \begin{tabular}{@{}c@{}} +3.98 \\ +3.10 \\ -0.50 \end{tabular}  \\
\hline
Brightness & \begin{tabular}{@{}c@{}} +0.85 \\ +0.78 \\ +0.19 \end{tabular}
& \begin{tabular}{@{}c@{}} +5.73 \\ +2.23 \\ +1.28 \end{tabular}
& \begin{tabular}{@{}c@{}} +0.25 \\ +5.02 \\ +1.76 \end{tabular}
& \begin{tabular}{@{}c@{}} +2.81 \\ +3.10 \\ -0.41 \end{tabular}  \\
\hline
Color Reduction & \begin{tabular}{@{}c@{}} +0.84 \\ +0.78 \\ +0.20 \end{tabular}
& \begin{tabular}{@{}c@{}} +5.72 \\ +2.23 \\ +1.36 \end{tabular}
& \begin{tabular}{@{}c@{}} +0.26 \\ +5.78 \\ +1.63 \end{tabular}
& \begin{tabular}{@{}c@{}} +3.97 \\ +3.10 \\ -0.41 \end{tabular}  \\
\hline
Gaussian Noise & \begin{tabular}{@{}c@{}} +0.84 \\ +0.84 \\ +0.22 \end{tabular}
& \begin{tabular}{@{}c@{}} +5.68 \\ +2.37 \\ +1.74 \end{tabular}
& \begin{tabular}{@{}c@{}} +4.79 \\ +3.48 \\ -0.47 \end{tabular}
& \begin{tabular}{@{}c@{}} +4.14 \\ +2.8 \\ -0.49 \end{tabular}  \\
\hline
\end{tabular}
\end{table}

\subsection{Comparison with state-of-the-art defenses} \label{s:comparison-soa}

To the best of our records, the detection method presented in this paper is the first attempt to defend CNNs against attacks from robust AEs in a white-box setting with input transformations. Previous work that used input transformations to detect white-box AEs did not consider robust AEs.
This is the case of two works: Xie et al.~\cite{xie2017mitigating} and Prakash et al.~\cite{prakash}, which used randomization and pixel deflection, respectively. Randomization\footnote{2017 NIPS adversarial defense competition runner-up --- code available at \url{https://github.com/anishathalye/obfuscated-gradients/tree/master/randomization}} consists in concatenating two input transformations, namely rescaling and padding; pixel deflection is an image transformation that swaps nearby pixel values. We compared two versions: the white-box version (naive), and the full defense with Wavelet Denoiser\footnote{The code for this part was taken from \url{https://github.com/anishathalye/pixel-deflection}}.

Furthermore, our comparison also considered VisionGuard \cite{visionguard}, which uses an architecture similar to \BASELINE, with JPEG compression as input transformation (which is not differentiable).

For a fair comparison, the evaluation was carried out on a single network (Inception-v4), as the existing methods did not consider multi-network architectures. The transformation used in our method is Gaussian noise, as it proved to be the best-performing among the evaluated ones. The evaluation metric is a tuple representing the true positive and false positive rates. This choice is motivated by the fact that pixel deflection and randomization are not merely detection methods but defenses (i.e., they modify the input image to correctly classify it). The true and false positive rates for the two detection methods (ours and VisionGuard), are computed by just selecting the threshold that provides the best rate (instead of computing the AUC of the ROC graph).

The results are summarized in Table \ref{t:comparison}. As it can be noted from the table, our method exhibits comparable performance with repspect the other methods for the detection of standard AEs (L-BFGS and CW columns). On the contrary, none of the other methods is capable of properly detecting robust AEs ($M_{\text{test}}$ column), while our method provides a high detection performance. Please note that the AEs in $M_{\text{test}}$ 
have not been generated to be robust to 
pixel deflection (with Wavelet denoiser) and VisionGuard (i.e., they have been generated to be robust to translation, blur, and Gaussian noise), which are even based on non-differentiable functions.
This suggests that there exists a sort of transferability effect for robust AEs: future work will investigate in this direction.

\begin{table}[]
\centering
\caption{Comparison with state-of-the-art methods using input transformations against the same threat model. The two values in each cell denote the true positive and false positive rates, respectively.}
\label{t:comparison}
\begin{tabular}{|l|c|c|c|}
\hline
Method                & L-BFGS & CW & $M_{\text{test}}$ \\ \hline
Randomization \cite{xie2017mitigating}         & \begin{tabular}{@{}c@{}} 0.983 \\ 0.085 \end{tabular}   & \begin{tabular}{@{}c@{}} 0.986 \\ 0.088 \end{tabular} &  \begin{tabular}{@{}c@{}} 0.010 \\ 0.070 \end{tabular}  \\ \hline
\begin{tabular}{@{}lc@{}} Pixel Deflection \cite{prakash} \\ (naive)  \end{tabular}       & \begin{tabular}{@{}c@{}} 0.948 \\ 0.083 \end{tabular}   & \begin{tabular}{@{}c@{}} 0.982 \\ 0.082 \end{tabular} & \begin{tabular}{@{}c@{}} 0.004 \\ 0.071 \end{tabular}   \\ \hline
\begin{tabular}{@{}lc@{}} Pixel Deflection \cite{prakash} \\ (w/ Wavelet denoiser)  \end{tabular} & \begin{tabular}{@{}c@{}} 0.985 \\ 0.125 \end{tabular}   & \begin{tabular}{@{}c@{}} 0.986 \\ 0.123 \end{tabular} &  \begin{tabular}{@{}c@{}} 0.048 \\ 0.111 \end{tabular}  \\ \hline
\begin{tabular}{@{}lc@{}}VisionGuard \cite{visionguard} \\ (JPEG92)  \end{tabular} & \begin{tabular}{@{}c@{}} \textbf{0.997} \\ 0.060 \end{tabular}   & \begin{tabular}{@{}c@{}} \textbf{0.992} \\ 0.078 \end{tabular} &  \begin{tabular}{@{}c@{}} 0.043 \\ 0.149 \end{tabular}  \\ \hline
\begin{tabular}{@{}lc@{}} Ours \\ (Gaussian noise)  \end{tabular} & \begin{tabular}{@{}c@{}} 0.978 \\ 0.088 \end{tabular}   & \begin{tabular}{@{}c@{}} 0.954 \\ 0.119 \end{tabular} &  \begin{tabular}{@{}c@{}} \textbf{0.944} \\ 0.080 \end{tabular} \\ \hline
\end{tabular}
\end{table}

\subsection{Discussion and future work}
In the light of the above experimental results, some aspects are worth to be discussed.
\paragraph{Best transformation} among the transformations that
were tested, some performed better than others. For example, rotation shows poor performance for all kinds of adversarial datasets. Others, such as contrast and brightness changes, and bit depth color reduction, show poor performance in detecting robust AEs, even though the AEs were not specifically robust to those transformations.
Blur is not among the best-performing either, while scale, translation, shear, mirror and Gaussian noise consistently show good performance.

In general, by the results of Section~\ref{s:comparison-soa} it emerges that to use our method with the \ENHANCED 
architecture 
one has to accept slightly lower performance in detecting standard AEs to be effective against all types of AEs. This happens because it has been experimentally found that the 
defense perturbation makes standard AEs robust to the input transformations considered in this work.
%

\paragraph{Results cannot be generalized to any CNN} it is not possible to state that translation will perform well as an AEs detector for any CNN, trained on any dataset. Performance will likely depend not only on the architecture and the dataset used, but also on the data augmentation techniques used during training. Further experimental evaluations should be performed in this direction.

Nevertheless, the analysis presented in this paper showed that networks with different architecture, but trained on the same dataset\footnote{ImageNet was chosen to provide real-world images; synthetic datasets as MNIST will surely show different results.}, present similar results for this kind of AEs detection system: the best and the worst performing transformations are the same, and in general the patterns of the detection performance of the transformations as functions of their parameters are similar.

\paragraph{Robust AEs are robust to non-differentiable transformations too}
The reason why robust AEs are not detected with Randomization or naive Pixel Deflection is evident: all those kinds of differentiable transformations are fooled by robust AEs because of the similarity of the transformations with respect to the ones used to craft them.
However, the reasons why VisionGuard and Pixel Deflection with Wavelet Denoiser are not able to detect robust AEs is still unclear and it will surely worth investigations. The input transformations used by these works are JPEG compression and Wavelet denoiser, which are non-differentiable input transformations and, therefore, cannot be used to generate robust AEs, at least as defined in this work. Our hypothesis (to be validated by future experiments) is that there exists some kind of transferability of AEs not only between different architectures of neural networks, but also in robustness between transformations.

\paragraph{Open issues}
The following aspects require further investigation to be generalized.
\begin{itemize}
  \item Due to space limits, not all the most common kinds of AEs were considered in the evaluation. Although the results of this experimental evaluation cannot be generalized to the entire spectrum of AEs, this work can be considered as a starting point to exhaustively test the detection potentiality of input transformations.
The loss function used to optimize the robust AEs is the cross-entropy loss. Other losses would certainly lead to different results. However, the process for generating the defense perturbation is general enough to account for robust AEs crafted with different loss functions, which will be explored in a future work.
  \item This paper considered a detection system using threshold-based binary classification. By varying the threshold it is possible to plot ROC graphs that help understand which transformations are best suited for this kind of detection. However, at run-time a certain threshold must be chosen (one for each network). The thresholds should take into account all the possible kinds of AEs that one wants to detect, in order to average the performance over the entire possible range of different AEs.
  \item The reason why non-differentiable input transformations are not able to detect robust AEs is still not clear and should be investigated in details.
\end{itemize}


\section{Conclusions} \label{s:concl}
This paper introduced a methodology to detect adversarial examples for CNNs. The method exploits the detection power of input image transformations for standard adversarial examples, which have been extensively tested to discover those transformations that are more suitable for this kind of detection problem.

Although robust adversarial examples can significantly degrade the performance of simple detection architectures (\BASELINE and \VOTINGBASELINE), this paper presented a counter-measure against robust adversarial examples based on the generation of a defense perturbation. This perturbation allows making robust adversarial examples sensitive again to input transformations and it can be used to achieve very good detection performance for both standard and robust adversarial examples.
Majority voting for multi-CNN systems has also been introduced to further improve the detection performance.


Future work will investigate extensions of the approaches presented in this paper to understand whether different kinds of attacks can fool the proposed detection systems. Also, further tests will be conducted to clarify the role of the data distribution used to generate the defense perturbation, to better comprehend whether it is possible for an attacker to fool the proposed detection systems without knowing the exact data distribution used to craft the defense.

\bibliographystyle{IEEEtran}
\bibliography{references}

\end{document}